%% file: main.tex
\documentclass[letterpaper, 10 pt, conference]{ieeeconf}  

\IEEEoverridecommandlockouts                              

\usepackage{graphicx} 
\usepackage[backend=biber,style=ieee,mincitenames=1,maxcitenames=2,natbib=true]{biblatex}
\let\labelindent\relax 
\usepackage{support-caption}
\usepackage{subcaption}
\DeclareCaptionLabelSeparator{periodspace}{.\quad}
\usepackage[inline]{enumitem}
\usepackage{amsmath} 
\usepackage[hidelinks]{hyperref}
\usepackage[capitalise,nameinlink]{cleveref}
\usepackage{amssymb}
\usepackage{doi}
\usepackage{xcolor}
\usepackage{siunitx}
\usepackage[super]{nth}

\newcommand{\paraDraft}[1]{\ifdefined\draft\subsubsection*{\color{orange}\textbf{#1}}\fi}
\newcommand{\paraFinal}[1]{\subsubsection*{\textbf{#1}}}

\ifdefined\draftTikz
    \usepackage{tikz}
    \usetikzlibrary{arrows,calc,decorations.markings,intersections,plotmarks,shapes}
    \usepackage{pgfplots}
    \usepackage{pgfplotstable}
    \usepgfplotslibrary{fillbetween}
    \pgfplotsset{compat=newest}
    \usetikzlibrary{external}
    \tikzset{%
        font=\footnotesize,
        external/up to date check=md5,
    }
    \pgfplotsset{%
        tick align=outside,
        tick pos=left,
        major tick length=2pt,
        xlabel shift = -3 pt,
        ylabel shift = -3 pt,
        every axis plot/.append style={%
            font=\footnotesize,
        }, 
        legend cell align={left},
        every axis legend/.append style={%
            font=\footnotesize,
        },
        every tick label/.append style={%
            font=\footnotesize,
        },
    }
\fi
\usepackage[acronym]{glossaries}
\newacronym{to}{TO}{Trajectory Optimization}
\newacronym{mpc}{MPC}{Model Predictive Controller}
\newacronym{mpc2}{MPC}{Model Predictive Control}
\newacronym{mpcc}{MPCC}{Mathematical Program with Complementarity Constraints}
\newacronym{miqp}{MIQP}{Mixed Integer Quadratic Programming}
\newacronym{minlp}{MINLP}{Mixed Integer Non-Linear Program}
\newacronym{th}{TH}{time horizon}

\title{\LARGE \bf
    Non-prehensile Planar Manipulation via Trajectory Optimization with Complementarity Constraints
}

\author{Jo\~{a}o Moura$^{1,2}$, Theodoros Stouraitis$^{1}$, and Sethu Vijayakumar$^{1,2}$
    \thanks{$^{1}$ Authors are with the School of Informatics, The University of Edinburgh, Edinburgh, U.K.}%
    \thanks{$^{2}$ Authors are with The Alan Turing Institute, London, U.K.}%
}

\begin{document}

\maketitle
\thispagestyle{empty}
\pagestyle{empty}

\begin{abstract}
Contact adaptation is an essential capability when manipulating objects.
Two key contact modes of non-prehensile manipulation are sticking and sliding.
This paper presents a~\acrfull*{to} method formulated as a~\acrfull*{mpcc}, which is able to switch between these two modes.
We show that this formulation can be applicable to both planning and~\acrfull*{mpc2} for planar manipulation tasks.
We numerically compare:
(i) our planner against a mixed integer alternative, showing that the~\acrshort*{mpcc} planner converges faster, scales better with respect to the~\acrfull*{th}, and can handle environments with obstacles;
(ii) our controller against a state-of-the-art mixed integer approach, showing that the~\acrshort*{mpcc} controller achieves improved tracking and more consistent computation times.
Additionally, we experimentally validate both our planner and controller with the KUKA LWR robot on a range of planar manipulation tasks.
See our accompanying video here:~\href{https://youtu.be/EkU6YHMhjto}{\color{blue!50!black}https://youtu.be/EkU6YHMhjto}.
\end{abstract}


\section{Introduction}\label{sec:introduction}
\input{sections/introduction.tex}

\section{Background}\label{sec:background}
\input{sections/background.tex}

\section{Method}\label{sec:method}
\input{sections/method.tex}

\section{Experiments and Results}\label{sec:experiments}
\input{sections/experiments.tex}

\section{Summary and Discussion}\label{sec:conclusion}
\input{sections/conclusion.tex}






\section*{ACKNOWLEDGMENT}

This research is supported by the EU H2020 projects Enhancing Healthcare with Assistive Robotic Mobile Manipulation (HARMONY, 101017008) and Memory of Motion (MEMMO, 780684), and the Kawada Robotics Corporation.




\clearpage

\printbibliography

\end{document}

%% file: sections/introduction.tex
\paraDraft{Contextualization}
Moving beyond the typical pick-and-place tasks, towards  non-prehensile manipulation, requires providing robots with the capability of adapting contact locations on the object.
However, achieving reliable robot manipulation via contact adaptation still poses many challenges due to, among other factors:
\begin{enumerate*}[label=(\alph*)]
    \item the under-actuated and hybrid nature of the problem~\cite{Mason1986,Hogan2020}; and
    \item the uncertainties arising from the frictional contact interactions~\cite{Zhou2017b,Bauza2017}.
\end{enumerate*}
Recent works~\cite{Hogan2017,Toussaint2019,Stouraitis2020b} have been addressing the problem of contact adaptation by developing control and motion planning methods based on~\acrfull*{to} that explicitly incorporate models of the contact interaction.
Nevertheless, there are still many open questions on both making plans with complex contact interactions~\cite{Toussaint2020} realizable by robots
and expanding the capabilities of current contact-aware controllers~\cite{Hogan2017,Aydinoglu2019} to reliably handle more challenging environments.

\subsection{Related Work}\label{subsec:related_work}

\begin{figure}[t]
    \centering
    \includegraphics[width=0.98\linewidth]{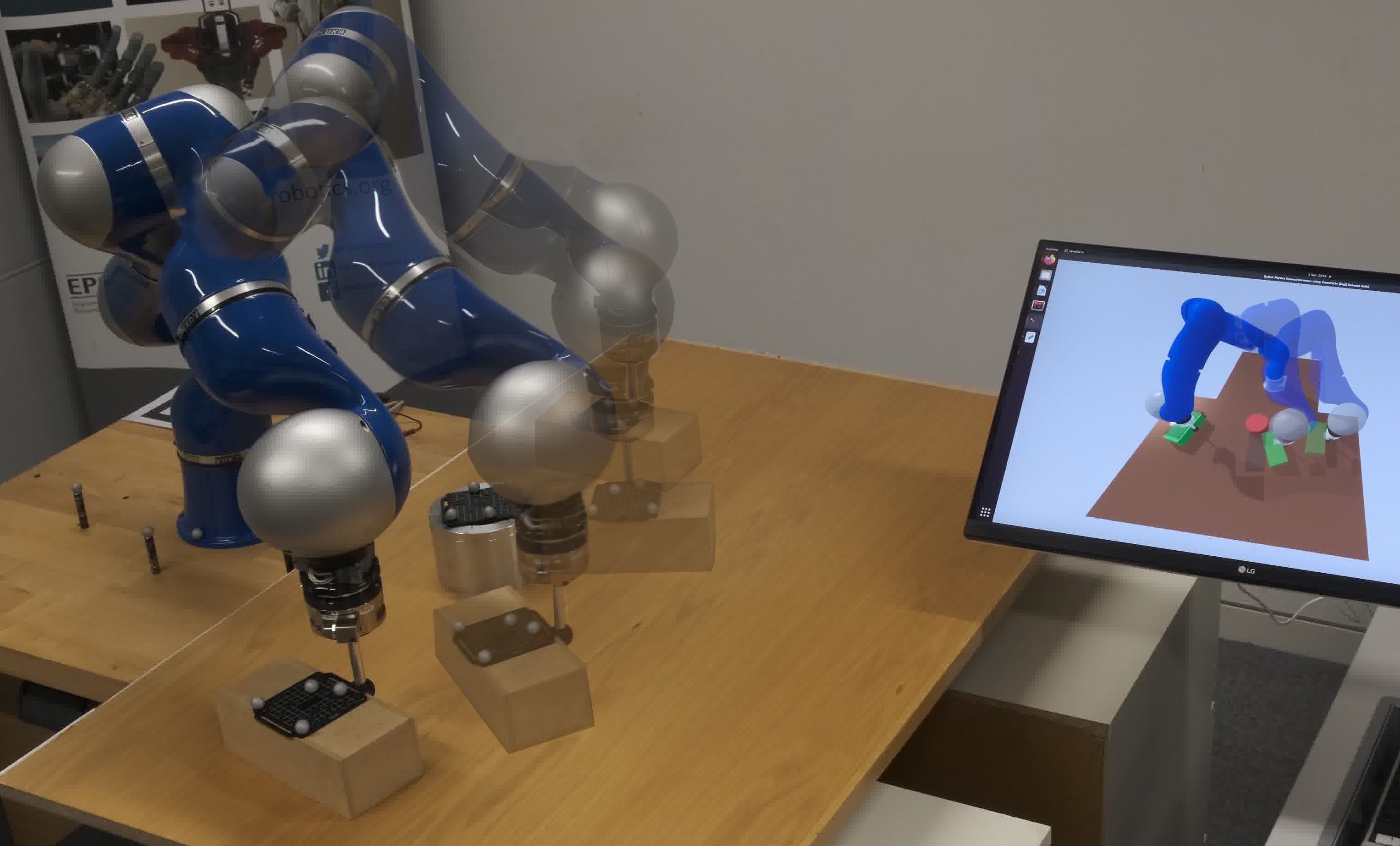}
    \caption{Experimental setup where the robot plans and controls the motion of an object to push it to the target, via sliding contact and while avoiding an obstacle.}
    \label{fig:teaser}
\end{figure}

\paraDraft{Non-prehensile pusher-slider example}
Non-prehensile manipulation, a term introduced by \citet{Mason1999}, refers to manipulation without grasping,~\textit{i.e.} the relative pose between the object and the robots’ end-effectors can change throughout the interaction.
While that type of manipulation can allow robots to execute a wider range of tasks~\cite{Mason1986}, executing such tasks with robots can prove itself quite challenging, in part due to various mismatches between the physical world and the respective models used for planning and control.
One prominent and widely used task, also introduced by~\citet{Mason1986},  
for studying non-prehensile manipulation is the planar pusher-slider example.
This example, as shown in Fig.~\ref{fig:teaser}, consists of a flat object---the slider---moving on a planar surface, pushed by the end-effector of the robot---the pusher.
The pusher-slider example is especially useful for exploring concepts such as the limit-surface model~\cite{Goyal1989}, the motion cone concept~\cite{Mason1986}, its generalization to a broader set of planar tasks~\cite{Chavan2018}, tactile feedback~\cite{Lynch1992}, and dynamics learning~\cite{Zhou2016,Bauza2017}.
Recently,~\citet{Hogan2020} proposed a~\acrfull*{mpc} for reactive tracking of nominal paths while reacting to disturbances.
They incorporate the selection of different contact modes, such as sticking and sliding contact, in the~\acrshort*{mpc} optimization by using a mixed integer formulation.
In this work, we investigate the limitations of this formulation in the context of both control and planning of planar sliding motions.

\paraDraft{Complementarity in rigid-body optimization}
Recent works~\cite{Toussaint2019,Stouraitis2018,Onol2019,Sleiman2019} have developed~\acrshort*{to} methods for planning robot manipulation tasks requiring contact changes.
The underlying formulation of these works follows one of two classes: {\it contact-implicit}~\cite{Posa2014} or {\it multi-mode}~\cite{Toussaint2019,Stouraitis2020}.
Here, we focus on the former which expresses the hybrid nature of the contact change as a~\acrfull*{mpcc}.
For the problem of making and breaking of contact, the complementarity constraint typically takes the form of~$0 \leq d \bot f \geq 0$, where~\(d\) and \(f\) are, respectively, the distance and the normal force between two objects of interest.
This constraint enforces both unilateral forces and that there is no penetration, while encoding the hybrid condition that the objects are either in contact or apart having, respectively, zero distance or zero contact force between them. 
To the extend of our knowledge no work has demonstrated the application of~\acrshort*{mpcc} for robot control problems, due to its computational requirements, yet we show that it is a particularly well suited formulation for both planning and control of planar non-prehensile manipulation problems with sticking and sliding contacts.

\subsection{Problem Statement}

\paraDraft{Pose research question}
In the~\nameref{subsec:related_work} we discuss a control formulation that explicitly handles switching contact modes, which requires high level methods to provide nominal paths.
There are a number of~\acrshort*{to} works capable of generating those nominal hybrid motion paths,~\textit{i.e.} involving change of contacts, however, realizing these on hardware is still a subject of research.
Hence, we ask ourselves~\textit{"what is an appropriate numerical optimal control formulation for non-prehensile manipulation problems with sliding contact that is applicable to both control and planning problems?"}
As a typical example scenario, consider a robot pushing an object to a goal, as shown in~\cref{fig:teaser}.
In such a case, the robot needs to address the following challenges: 
\begin{itemize}
    \item It has to plan a trajectory using both sticking and sliding contact modes to drive the object to the target while avoiding the obstacle;
    \item It has to control the motion of the object to track the planned trajectory under the uncertainties of the frictional contact interactions and other unknown disturbances.
\end{itemize}

\paraFinal{Contributions}
This work addresses these challenges, by:
\begin{itemize}
    \item Proposing a single~\acrshort*{mpcc} formulation for both planning and control (\acrshort*{mpc}) that enables switching between sticking and sliding contact modes;
    \item Comparing it numerically with an alternative mixed integer formulation, showing that the~\acrshort*{mpcc} achieves
      \begin{enumerate*}[label=\roman*)]
        \item smaller tracking errors,
        \item more reliable computation times under large disturbances,
        \item and better scalability with regards to the~\acrfull*{th} and to a broader range of problems,~\textit{e.g.} problems with obstacles;
      \end{enumerate*}
    \item And finally, experimentally validating our~\acrshort*{mpcc} formulation using the KUKA LWR robot hardware.
\end{itemize}

%% file: sections/background.tex
\paraDraft{Motion model building block}

This section revises three key models/assumptions 
commonly used in the literature for modelling the planar pusher-slider system.
Namely, the quasi-static assumption, the limit surface, and the friction cone.

\subsection{System Description}

\begin{figure}[t]
    \centering
    \ifdefined\draftTikz
        \tikzsetnextfilename{fig_single_contact_diag}
        \input{Tikz/fig_single_contact_diag.tex}
    \else
        \includegraphics{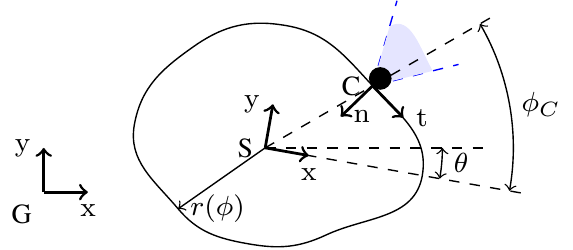}
    \fi
    \caption{Illustration of the pusher-slider system.}
    \label{fig:single_contact_diag}
\end{figure}

\paraDraft{Describe system illustration}
Fig.~\ref{fig:single_contact_diag} illustrates a top down view of the planar pusher-slider system, where $(^G\mathrm{x}_S,^G\mathrm{y}_S)$ and~$\theta$ are, respectively, the Cartesian position and orientation of the sliding object---the slider---with respect to a global reference~$G$. 
Given a known geometry of the slider~$r(\phi)$, a single parameter~$\phi_C$ is sufficient to compute the position of the pushing object---the pusher---as well as the contact point~$C$, normal~$n$ and tangential~$t$ directions.

\subsection{Quasi-static Assumption}

\paraDraft{Quasi-static simplification}
Let~$\omega=\begin{bmatrix}v_\mathrm{x},v_\mathrm{y},\omega_\mathrm{z}\end{bmatrix}^\top$ and~$\tau=\begin{bmatrix}f_\mathrm{x},f_\mathrm{y},\tau_\mathrm{z}\end{bmatrix}^\top$ be, respectively, the vectors of generalized velocities and forces for the slider.
Then we can write its Newtonian dynamics as
\begin{equation}
    I\dot{\omega}=\tau_C+\tau_P,
    \label{eq:rigid_body_dynamics}
\end{equation}
where~$I$ is the inertia matrix of the slider,~$\tau_C$ are the forces due to lateral contacts (pusher) and~$\tau_P$ are the forces due to the sliding friction with the planar surface.
However, it is common practice in the literature to simplify the equation of motion~\eqref{eq:rigid_body_dynamics}, by neglecting the inertial forces~$I\dot{\omega}$, based on the observation that for low velocities~$\omega$, frictional contact forces dominate the motion dynamics of the pusher-slider system~\cite{Mason1986}.
Thus, we obtain~$\tau_C=-\tau_P$.

\subsection{Ellipsoidal Approximation of the Limit Surface}

\paraDraft{Limit surface approximation}
Now, we need to find a model for computing the friction forces~$\tau_P$.
\citet{Goyal1989} introduced the idea of limit surface that describes the mapping between~$\tau_P$ and~$\omega$.
In this work, we use the following ellipsoidal approximation to model the limit surface
\begin{equation}
    H(\tau_P) \triangleq \tau_P^\top L\tau_P,
    \label{eq:limit_surface_ellipse}
\end{equation}
where~$L$ is a positive definite matrix~\cite{Zhou2016}.
This model captures well the shape of the limit surface for pusher-slider interactions with uniform pressure distributions~\cite{Howe1996}.
Given a convex limit surface~$H(\tau_P)$, the slider velocities will be perpendicular to the limit surface~\cite{Hogan2020}, resulting in
\begin{equation}
    \omega=\nabla H(\tau_P)=L\tau_P.
    \label{eq:limit_surface_ellipse_result}
\end{equation}

\subsection{Motion Model}

\paraDraft{State and action description}
This subsection describes the motion model used 
to predict and optimize the motion of the pusher-slider system.
Given the quasi-static assumption, the motion model becomes a geometric kinematic model and, hence, the state simply corresponds to the system configuration
\begin{equation}
    x =
    \begin{bmatrix}
        ^G\mathrm{x}_S,
        ^G\mathrm{y}_S,
        \theta,
        \phi_C
    \end{bmatrix}^\top,
\end{equation}
which describes the position and orientation of the pusher and slider.
A possible representation for the control is
\begin{equation}
    u =
    \begin{bmatrix}
        f_n,
        f_t,
        \dot{\phi}_{C+},
        \dot{\phi}_{C-}
    \end{bmatrix}^\top,
    \label{eq:controls}
\end{equation}
where~$f_n$ and~$f_t$ are, respectively, the normal and tangential forces applied by the pusher to the slider, and~$\dot{\phi}_C=\dot{\phi}_{C+}-\dot{\phi}_{C-}$ is the angular rate of sliding.
Note that~\eqref{eq:controls} is a redundant representation that we will discuss in the next subsection.

\paraDraft{Dynamics description}
For single contact and a rectangular object with contact~$C$ and local~$S$ frames aligned, through simple geometric transformations, using the ellipsoidal approximation of the limit surface~\eqref{eq:limit_surface_ellipse_result}, and the quasi-static assumption, we obtain the following equations of motion for the pusher-slider system~\cite{Hogan2020}
\begin{equation}
    \dot{x} = f(x,u) = 
    \begin{bmatrix}
        RLJ_C^\top & 0 & 0\\
        0          & 1 & -1
    \end{bmatrix}u,
    \label{eq:system_dynamics}
\end{equation}
with~$R(\theta)$ being the~$\mathrm{xy}$-plane rotation matrix between the slider local frame~$S$ and the global frame~$G$, and
\begin{equation*}
    J_C =
    \begin{bmatrix}
        1&0&-^S\mathrm{y}_C\\
        0&1&^S\mathrm{x}_C
    \end{bmatrix}
\end{equation*}
being the contact Jacobian matrix.

\subsection{Friction Constraints}

\paraDraft{Hybrid dynamics}
Depending on the values that the redundant set of controls in~\eqref{eq:controls} take, the pusher-slider system can exhibit different dynamic behaviours.
Those different behaviours, or dynamic modes, include sticking or sliding contact between the pusher and the slider, or even no contact at all.
Originally,~\citet{Hogan2020b} proposed a hybrid dynamics model for the equations of motion~$f(\cdot)$.
Later, they~\cite{Hogan2020} expressed the hybrid nature of the system via constraints on the controls~\eqref{eq:controls}.
In that way, they separate the continuous dynamics of the system, as in~\eqref{eq:system_dynamics}, from the hybrid component corresponding to different active contact modes.
The advantage of this approach, in the context of formulating an optimization, is the ease of transcribing the selection of contact modes, hence, we stick with this approach. 

\begin{figure}[h]
    \begin{center}
        \begin{subfigure}{.15\textwidth}
            \centering
            \ifdefined\draftTikz
                \tikzsetnextfilename{fig_mode_sticking}
                \input{Tikz/fig_mode_sticking.tex}
            \else
                \includegraphics{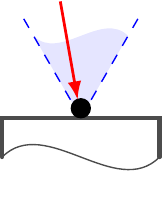}
            \fi
            \vspace{-5mm}
            \caption{}
            \label{fig:contact_mode_sticking}
        \end{subfigure}
        \begin{subfigure}{.15\textwidth}
            \centering
            \ifdefined\draftTikz
                \tikzsetnextfilename{fig_mode_sliding_counterclockwise}
                \input{Tikz/fig_mode_sliding_counterclockwise.tex}
            \else
                \includegraphics{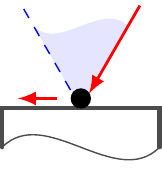}
            \fi
            \vspace{-5mm}
            \caption{}
            \label{fig:contact_mode_sliding_counterclockwise}
        \end{subfigure}
        \begin{subfigure}{.15\textwidth}
            \centering
            \ifdefined\draftTikz
                \tikzsetnextfilename{fig_mode_sliding_clockwise}
                \input{Tikz/fig_mode_sliding_clockwise.tex}
            \else
                \includegraphics{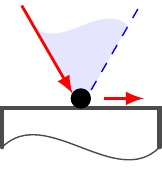}
            \fi
            \vspace{-5mm}
            \caption{}
            \label{fig:contact_mode_sliding_clockwise}
        \end{subfigure}
        \vspace{2mm}
        \caption{
            Slider contact modes:
            \protect(\subref{fig:contact_mode_sticking}) sticking;
            \protect(\subref{fig:contact_mode_sliding_counterclockwise}) sliding~\textbf{c}ounter\textbf{c}lock\textbf{w}ise (ccw);
            \protect(\subref{fig:contact_mode_sliding_clockwise}) sliding~\textbf{c}lock\textbf{w}ise (cw).
        }
        \label{fig:contact_modes}
    \end{center}
\end{figure}

\paraDraft{Contact modes}
Regardless of the active contact mode, we enforce a unilateral constraint that allows only pushing of the slider, as~$\mathcal{U}_0: f_n\geq0$.
Fig.~\ref{fig:contact_modes} illustrates the three contact modes considered, corresponding to the following sets of constraints
\begin{equation*}
    \mathcal{U}_1:
    \begin{cases}
        \dot{\phi}_C=0\\
        |f_t|\leq\mu f_n
    \end{cases}\hspace{-4.5mm},\;
    \mathcal{U}_2: 
    \begin{cases}
        \dot{\phi}_C>0\\
        f_t=\mu f_n
    \end{cases}\hspace{-4.5mm},\,\text{and}\;
    \mathcal{U}_3: 
    \begin{cases}
        \dot{\phi}_C<0\\
        f_t=-\mu f_n 
    \end{cases}\hspace{-4.5mm}.
\end{equation*}
For a sticking contact, corresponding to~$\mathcal{U}_1$ and illustrated by Fig.~\ref{fig:contact_mode_sticking}, the applied force has to remain within the friction cone and the angular rate of sliding~$\phi_C$ has to be zero.
Note that $\mu$ is the friction coefficient between the pusher and the slider.
For the sliding~\textbf{c}ounter\textbf{c}lock\textbf{w}ise (ccw) contact mode, corresponding to~$\mathcal{U}_2$ and illustrated by Fig.~\ref{fig:contact_mode_sliding_counterclockwise}, the applied force belongs to the one edge of the friction cone while~$\phi_C$ is positive.
Finally, for the~\textbf{c}lock\textbf{w}ise (cw) contact mode, corresponding to~$\mathcal{U}_3$ and illustrated by Fig.~\ref{fig:contact_mode_sliding_clockwise}, the applied force belongs to the other edge of the friction cone while~$\phi_C$ is negative.

%% file: Tikz/fig_single_contact_diag.tex
\begin{tikzpicture}
    \def\U{0.9cm}
    \def\thetaval{-10}
    \def\psival{40}
    \edef\psitheta{0}
    \pgfmathparse{\psival+\thetaval}
    \edef\psitheta{\pgfmathresult}
    \edef\angNorm{0}
    \pgfmathparse{\psival+\thetaval+14+180}
    \edef\angNorm{\pgfmathresult}
    \coordinate (origin) at (0,0);
    \coordinate (objectCentre) at ($(origin)+(2.5*\U,0.5*\U)$);
    \coordinate (objectB1) at ($(objectCentre)+(\thetaval+0:1.8*\U)$);
    \coordinate (objectB2) at ($(objectCentre)+(\thetaval+\psival:1.4*\U)$);
    \coordinate (objectContact) at (objectB2);
    \coordinate (objectB3) at ($(objectCentre)+(\thetaval+90:1.4*\U)$);
    \coordinate (objectB4) at ($(objectCentre)+(\thetaval+135:1.4*\U)$);
    \coordinate (objectB5) at ($(objectCentre)+(\thetaval+180:1.5*\U)$);
    \coordinate (objectB6) at ($(objectCentre)+(\thetaval+-135:1.2*\U)$);
    \coordinate (objectB7) at ($(objectCentre)+(\thetaval+-90:1.1*\U)$);
    \coordinate (objectB8) at ($(objectCentre)+(\thetaval+-45:1.2*\U)$);
    \draw[rotate=\thetaval] (objectB1) to [out=90,in=-45+8] (objectB2)
    to [out=135+8,in=0] (objectB3) to [out=180,in=45] (objectB4)
    to [out=-135,in=90] (objectB5) to [out=-90,in=135+5] (objectB6)
    to [out=-45+5,in=-180] (objectB7) to [out=0,in=-135-10] (objectB8) to [out=45-10,in=-90] cycle;
    \draw[dashed, blue] (objectContact) -- ($(objectContact)+(\angNorm+180-30:1.0*\U)$);
    \draw[dashed, blue] (objectContact) -- ($(objectContact)+(\angNorm+180+30:1.0*\U)$);
    \fill[blue!10] (objectContact) -- ($(objectContact)+(\angNorm+180-30:0.7*\U)$) to [out=120,in=30] ($(objectContact)+(\angNorm+180+30:0.7*\U)$) -- cycle;
    \def\dashL{3.0*\U}
    \draw[dashed] (objectCentre) -- ($(objectCentre)+(0:2.5*\U)$);
    \draw[dashed] (objectCentre) -- ($(objectCentre)+(\thetaval:\dashL)$);
    \draw[dashed] (objectCentre) -- ($(objectCentre)+(\psitheta:\dashL)$);
    \draw[<->] ($(objectCentre)+(0:2.0*\U)$) arc (0:\thetaval:2.0*\U) node[midway, right] {$\theta$};
    \draw[<->] ($(objectCentre)+(\thetaval:2.8*\U)$) arc (\thetaval:\psitheta:2.8*\U) node[midway, right] {$\phi_C$};
    \draw[->] (objectCentre) -- (objectB6) node[right] {$r(\phi)$};
    \draw[thick,->] (origin) -- ++(0.5*\U,0) node[black,below] {$\mathrm{x}$};
    \draw[thick,->] (origin) -- ++(0,0.5*\U) node[black,left] {$\mathrm{y}$};
    \node at (origin) [below left] {G};
    \draw[thick,->, rotate=\thetaval] (objectCentre) -- ++(0.5*\U,0) node[black, below] {$\mathrm{x}$};
    \draw[thick,->, rotate=\thetaval] (objectCentre) -- ++(0,0.5*\U) node[black, left] {$\mathrm{y}$};
    \node at (objectCentre) [left] {S};
    \draw[thick,->, rotate=\angNorm] (objectContact) -- ++(0.5*\U,0) node[black, right] {$\mathrm{n}$};
    \draw[thick,->, rotate=\angNorm] (objectContact) -- ++(0,0.5*\U) node[black, right] {$\mathrm{t}$};
    \node at (objectContact) [left] {C};
    \draw[fill=black] ($(objectContact)+(\angNorm+180:0.12*\U)$) circle (0.12*\U);
\end{tikzpicture}

%% file: Tikz/fig_mode_sticking.tex
\begin{tikzpicture}
    \def\U{0.8cm}
    \coordinate (origin) at (0,0);
    \coordinate (cornerRight) at ($(origin)+(1*\U,0)$);
    \coordinate (endRight) at ($(cornerRight)+(0,-0.5*\U)$);
    \coordinate (cornerLeft) at ($(origin)+(-1*\U,0)$);
    \coordinate (endLeft) at ($(cornerLeft)+(0,-0.5*\U)$);
    \coordinate (arrowStart) at ($(origin)+(100:1.5*\U)$);
    \draw[very thick, black!70, line cap=round] (endLeft) -- (cornerLeft) -- (cornerRight) -- (endRight);
    \draw[black!70] (endLeft) to [out=45,in=-135] (endRight);
    \fill[blue!10] (origin) -- ($(origin)+(120:1.2*\U)$) to [out=-45,in=135] ($(origin)+(60:1.2*\U)$) -- cycle;
    \draw[dashed, blue] (origin) -- ($(origin)+(120:1.5*\U)$);
    \draw[dashed, blue] (origin) -- ($(origin)+(60:1.5*\U)$);
    \draw[fill=black, name path=circle] ($(origin)+(90:0.12*\U)$) circle (0.12*\U);
    \path[name path=S--O] (arrowStart) -- (origin);
    \path[name intersections={of=circle and S--O,by={A,B}}];
    \draw[red, -latex, thick] (arrowStart) -- (A);
\end{tikzpicture}

%% file: Tikz/fig_mode_sliding_counterclockwise.tex
\begin{tikzpicture}
    \def\U{0.8cm}
    \coordinate (origin) at (0,0);
    \coordinate (cornerRight) at ($(origin)+(1*\U,0)$);
    \coordinate (endRight) at ($(cornerRight)+(0,-0.5*\U)$);
    \coordinate (cornerLeft) at ($(origin)+(-1*\U,0)$);
    \coordinate (endLeft) at ($(cornerLeft)+(0,-0.5*\U)$);
    \coordinate (arrowStart) at ($(origin)+(60:1.5*\U)$);
    \draw[very thick, black!70, line cap=round] (endLeft) -- (cornerLeft) -- (cornerRight) -- (endRight);
    \draw[black!70] (endLeft) to [out=45,in=-135] (endRight);
    \fill[blue!10] (origin) -- ($(origin)+(120:1.2*\U)$) to [out=-45,in=135] ($(origin)+(60:1.2*\U)$) -- cycle;
    \draw[dashed, blue] (origin) -- ($(origin)+(120:1.5*\U)$);
    \draw[fill=black, name path=circle] ($(origin)+(90:0.12*\U)$) circle (0.12*\U);
    \path[name path=S--O] (arrowStart) -- (origin);
    \path[name intersections={of=circle and S--O,by={A,B}}];
    \draw[red, -latex, thick] (arrowStart) -- (A);
    \draw[red, -latex, thick] ($(origin)+(-0.3*\U,0.12*\U)$) --++ (-0.5*\U,0);
\end{tikzpicture}

%% file: Tikz/fig_mode_sliding_clockwise.tex
\begin{tikzpicture}
    \def\U{0.8cm}
    \coordinate (origin) at (0,0);
    \coordinate (cornerRight) at ($(origin)+(1*\U,0)$);
    \coordinate (endRight) at ($(cornerRight)+(0,-0.5*\U)$);
    \coordinate (cornerLeft) at ($(origin)+(-1*\U,0)$);
    \coordinate (endLeft) at ($(cornerLeft)+(0,-0.5*\U)$);
    \coordinate (arrowStart) at ($(origin)+(120:1.5*\U)$);
    \draw[very thick, black!70, line cap=round] (endLeft) -- (cornerLeft) -- (cornerRight) -- (endRight);
    \draw[black!70] (endLeft) to [out=45,in=-135] (endRight);
    \fill[blue!10] (origin) -- ($(origin)+(120:1.2*\U)$) to [out=-45,in=135] ($(origin)+(60:1.2*\U)$) -- cycle;
    \draw[dashed, blue] (origin) -- ($(origin)+(60:1.5*\U)$);
    \draw[fill=black, name path=circle] ($(origin)+(90:0.12*\U)$) circle (0.12*\U);
    \path[name path=S--O] (arrowStart) -- (origin);
    \path[name intersections={of=circle and S--O,by={A,B}}];
    \draw[red, -latex, thick] (arrowStart) -- (A);
    \draw[red, -latex, thick] ($(origin)+(0.3*\U,0.12*\U)$) --++ (0.5*\U,0);
\end{tikzpicture}

%% file: sections/method.tex
\subsection{Mixed Integer Formulation}

\paraDraft{Introduce mixed integer approach}
When developing an optimal control formulation for computing trajectories for the pusher-slider system that is hybrid, a natural way of incorporating 
contact modes switching is to include one binary variable $z$ per contact mode defining what mode is active at any instance of time.
Following this, we can then formalize a~\acrfull*{to}~\acrfull*{minlp} as follows
\begin{subequations}
    \begin{align}
        &\min_{x_i,u_i,z_i} &\quad& C(x_{0:N},u_{0:N-1},z_{1:3,0:N-1})\label{eq:mi:cost}\\
        &\text{subject to} &      & x_{i+1}=x_i + \Delta t f(x_i, u_i),\,x_i\in\mathcal{X},\label{eq:mi:dyn}\\
        &                  &      & u_i\in\mathcal{U}_0\,(\text{unilaterality}),\label{eq:mi:unilaterality}\\
        &                  &      & u_i\in\mathcal{U}_1\quad\text{if}\,z_{1,i}=1\,(\text{sticking}),\label{eq:mi:sticking}\\
        &                  &      & u_i\in\mathcal{U}_2\quad\text{if}\,z_{2,i}=1\,(\text{sliding ccw}),\label{eq:mi:sliding_ccw}\\
        &                  &      & u_i\in\mathcal{U}_3\quad\text{if}\,z_{3,i}=1\,(\text{sliding cw}),\label{eq:mi:sliding_cw}\\
        &                  &      & z_{1,i}+z_{2,i}+z_{3,i}=1,\label{eq:constraint6}
    \end{align}
    \label{eq:minlp}%
\end{subequations}
where~$\mathcal{X}$ is the set of feasible states, including state bounds and obstacles, and $i$ denotes the index of the knot, i.e.\ the discretization points of the transcribed problem.
The cost function is~$C(x_{0:N},u_{0:N-1},z_{1:3,0:N-1})=\bar{x}_N^\top W_{x_N}\bar{x}_N + \sum_{i=0}^{N-1}(\bar{x}_{i+1}^\top W_x\bar{x}_{i+1} + \bar{u}_i^\top W_u\bar{u}_i + \sum_{j=1}^3w_{z}z_{j,i})$, where~$\bar{x}_i=x_i-x_i^\ast$ is the difference between the state~$x_i$ and the goal state~$x_i^\ast$ ,
We implement~\eqref{eq:mi:sticking},~\eqref{eq:mi:sliding_ccw}, and~\eqref{eq:mi:sliding_cw} through the big M formulation~\cite{Nemhauser1988}. 

\paraDraft{Contrast with MIQP}
The~\acrshort*{minlp} in~\eqref{eq:minlp} is inspired by the~\acrfull*{miqp} proposed by~\citet{Hogan2020}.
The key difference is that the formulation  in~\cite{Hogan2020} requires linearization along a given nominal path (states and actions).
The~\acrshort*{miqp} is a suitable formulation for control because quadratic programs are generally very fast to solve.
However, it is unsuitable for planning, i.e. generating trajectories, as it requires both nominal states and actions as reference, and mixed integer programs tend to scale badly due to the combinatorial expansion when exploring the solution space of the integer variables.

\subsection{Complementary Constraints}

\paraDraft{Motivate Complementarity Constraints}
Another way of expressing different modes in hybrid systems is to use complementarity constraints, which have recently been widely used in many~\acrshort*{to} based planning methods for robotics~\cite{Posa2014,Aydinoglu2019,Sleiman2019,Toussaint2020}.
The complementarity constraints remove the need of using integer variables in the optimization problem.
Particular to the pusher-slider problem, the key insight for exploring complementarity constraints instead of a mixed integer formulation is that, even though the dynamics are hybrid, the transition between the different modes is continuous, meaning that the system can transition from the currently active contact mode to any other contact mode, at any instant in time.
Note that for the scenario of making and breaking of contact~\cite{Posa2014,Sleiman2019,Toussaint2020}, when far away from contact it is physically infeasible to instantaneously transition to contact.
This additional guard condition ensuring that mode transitions only occur from certain state space regions is nonexistent in our problem.

\paraDraft{Introduce new variables}
Let us introduce a vector of two variables that correspond to the edge of the friction cone
\begin{equation}
    \lambda_v \triangleq
    \begin{bmatrix}
        \lambda_-, 
        \lambda_+
    \end{bmatrix}^\top
    \triangleq
    \begin{bmatrix}
        \mu f_n-f_t, 
        \mu f_n+f_t
    \end{bmatrix}^\top.
\end{equation}
It is straightforward to show that the second inequality in~$\mathcal{U}_1$ corresponds to having~$\lambda_-,\lambda_+\geq0$.
We also introduced the control variables~$\dot{\phi}_{C+},\dot{\phi}_{C-}$ in~\eqref{eq:controls} rather than using~$\dot{\phi}_C=\dot{\phi}_{C+}-\dot{\phi}_{C-}$, as in~\cite{Hogan2020}, to enable the formulation of the complementarity constraint.
In this way, we can obtain: the sticking contact mode when both~$\dot{\phi}_{C+},\dot{\phi}_{C-}=0$ and~$\lambda_-,\lambda_+\geq0$; the sliding ccw contact mode~$\mathcal{U}_2$ when both~$\dot{\phi}_{C-},\lambda_-=0$ and~$\dot{\phi}_{C+}\geq0$; and the sliding cw contact mode~$\mathcal{U}_3$ when both~$\dot{\phi}_{C+},\lambda_+=0$ and~$\dot{\phi}_{C-}\geq0$.

\paraDraft{Introduce the complementarity constraints}
By defining the vector~$\dot{\phi}_v\triangleq\begin{bmatrix}\dot{\phi}_{C+},\dot{\phi}_{C-}\end{bmatrix}^\top$, we can easily verify that the complementarity constraint condition~$\lambda_v^\top\dot{\phi}_v=0$, with $\dot{\phi}_{C+},\dot{\phi}_{C-},\lambda_-\lambda_+\geq0$, simultaneously satisfies all the constraints required by the three contact modes.
Therefore, we can define the complementarity constraints as
\begin{equation}
    \mathcal{U}_{cc}: 
    \begin{cases}
        \dot{\phi}_{C+},\dot{\phi}_{C-},\lambda_-\lambda_+\geq0\\
        \lambda_v^\top\dot{\phi}_v + \varepsilon=0
    \end{cases},
\end{equation}
where we introduce a slack variable~$\varepsilon$.
This variable will be zero when the contact mode constraints are fully satisfied.
We introduce this variable because it is well known in the literature that problems with complementarity constraints are in practise difficult to solve, usually requiring relaxations~\cite{Nurkanovic2020}.

\subsection{Non-Linear Programming (NLP)}

\paraDraft{Present optimization}
We can then formalize our~\acrshort*{to}~\acrfull*{mpcc} as follows
\begin{subequations}
    \begin{align}
        &\min_{x_i,u_i,\varepsilon_i} &\quad& C(x_{0:N},u_{0:N-1},\varepsilon_{0:N-1})\label{eq:nlp:cost}\\
        &\text{subject to} &      & x_{i+1}=x_i + \Delta t f(x_i, u_i),\,x_i\in\mathcal{X},\label{eq:nlp:dyn}\\
        &                  &      & u_i\in\mathcal{U}_0\,(\text{unilaterality}),\label{eq:nlp:unilaterality}\\
        &                  &      & u_i,\varepsilon_i\in\mathcal{U}_{cc}\,(\text{complementarity}),\label{eq:nlp:complementarity}
    \end{align}
    \label{eq:mpcc}%
\end{subequations}
where the cost function is~$C(x_{0:N},u_{0:N-1},\varepsilon_{0:N-1})=\bar{x}_N^\top W_{x_N}\bar{x}_N + \sum_{i=0}^{N-1}(\bar{x}_{i+1}^\top W_x\bar{x}_{i+1} + u_i^\top W_uu_i + w_{\varepsilon_i}\varepsilon_i^2)$.
Changing formulation~\eqref{eq:mpcc} from a tracking to a planning problem is simply a matter of specifying the full nominal state trajectory~$x_{0:N}^\ast$ or specifying the final state target~$x_N^\ast$, respectively.
Also, using~\eqref{eq:mpcc} for planning full trajectories or as a~\acrfull*{mpc} is also a matter of changing the horizon~$N$, such that for a small enough horizon we can solve~\eqref{eq:mpcc} online in a control loop.

%% file: sections/experiments.tex
This section presents numerical and robot experiments performed using the~\acrshort*{mpcc} formulation given in~\eqref{eq:mpcc}.
We compare the~\acrshort*{mpcc} formulation against MIQP~\cite{Hogan2020} for tracking trajectories, and against~\acrshort*{minlp} given in~\eqref{eq:minlp} for planning trajectories.
We implemented all optimization using~\texttt{CasADi} \cite{Andersson2018}.
We used the~\texttt{Knitro} solver~\cite{Byrd2006} for both the~\acrshort*{mpcc} and~\acrshort*{minlp} problems and the~\texttt{Gurobi} solver~\cite{gurobi} for the~\acrshort*{miqp}.
We ran all the computations in a 64-bit Intel 16-Core i9 3.60GHz workstation with 64GB RAM.

\paraDraft{Report constants}
For every experiment, the cost function gain matrices are~$W_x=\text{diag}(1, 1, 0.01, 0.001)$,~$W_u=10^{-2}\text{diag}(1, 1, 0, 0)$, and~$W_{x_N}=10W_x$.
Note that planning problems only have a single state target~$x_N^\ast$, hence we only use~$W_{x_N}$.
For the mixed integer formulations~$w_z=0$.
For the~\acrshort*{mpcc} formulation~$w_{\varepsilon_i}=50$ for the planning problems while for the tracking problems~$w_{\varepsilon_i}=50$ for the first knot reducing exponentially to~$0.1$ for the last knot.
The number of discretization steps of the MPC is 25 steps with~$\Delta t=1/25\,\si{\second}$, which results in total~\acrshort*{th} of $T=1\,\si{\second}$.
We use a friction coefficient of~$\mu=0.2$ for the numerical experiments and~$\mu=0.1$ for the robot experiments.
We compute~$L$ in~\eqref{eq:limit_surface_ellipse_result} according to~\cite{Hogan2020}.

\paraDraft{Explain setup}
For the actual robot experiments, the~\acrshort*{mpc} runs in a~$50\,\si{\hertz}$ control loop. 
In each~\acrshort*{mpc} loop, we measure the position and orientation of the slider through a~\texttt{Vicon} tracking system to generate the initial state~$x_0$.
We then run the optimization~\eqref{eq:mpcc} and only make use of the first state~$x_1$ of the solution trajectory, from which we compute the position of the pusher and, through standard inverse kinematics, the respective configuration of the robot.
The accompanying video shows the robot experiments detailed in this section.

\begin{figure*}[t]
    \centering
    \ifdefined\draftTikz
        \tikzsetnextfilename{fig_trajectory_plots}
        \input{Tikz/fig_trajectory_plots.tex}
        \vspace{-3mm}
    \else
        \includegraphics{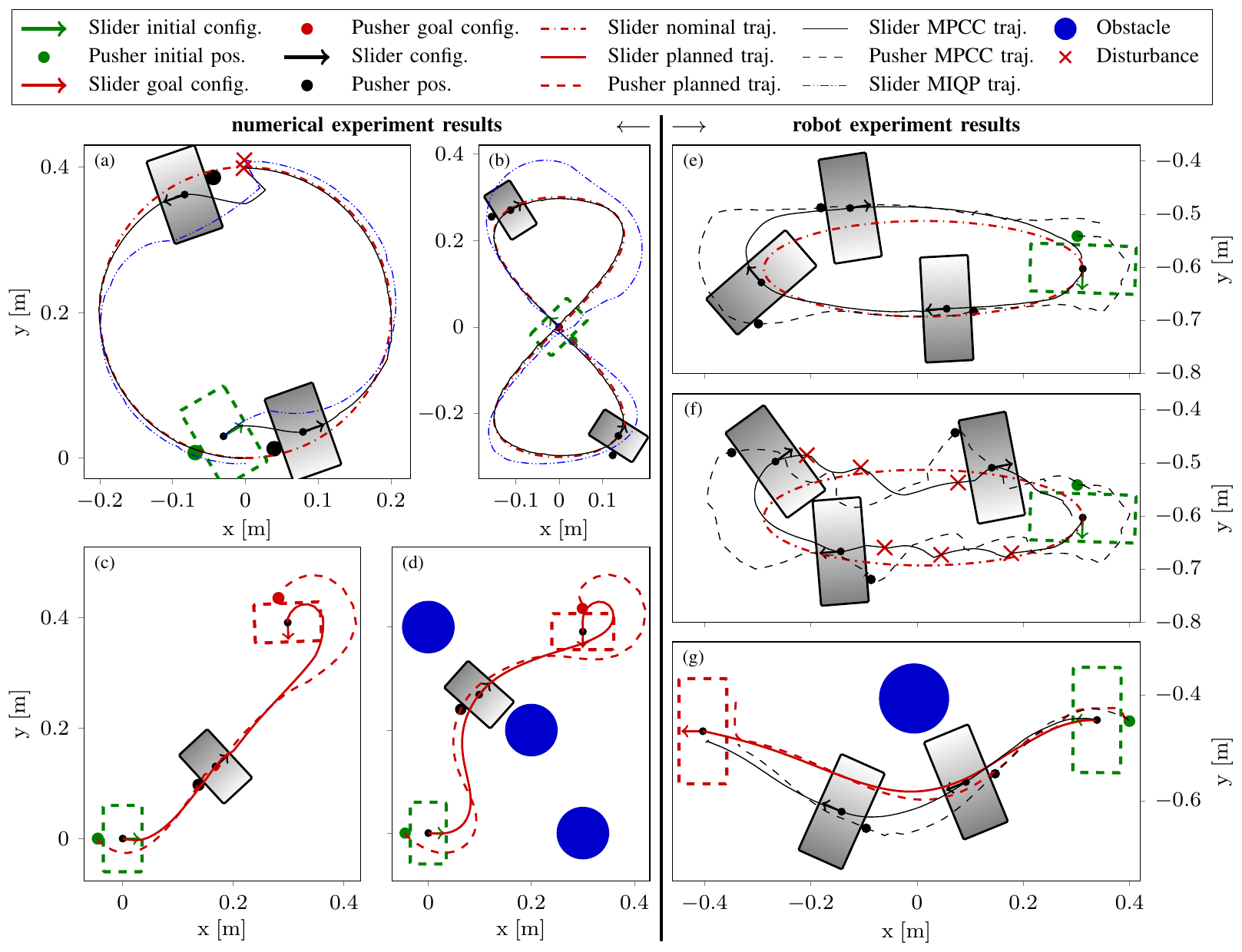}
    \fi
    \ifx\draftTikz\undefined
        \captionsetup[subfigure]{labelformat=empty}
        \subcaption{}\label{fig:tracking_circle_traj}
        \subcaption{}\label{fig:tracking_eight_traj}
        \subcaption{}\label{fig:plannning_without_obstacles}
        \subcaption{}\label{fig:planning_with_obstacles}
        \subcaption{}\label{fig:tracking_ellipse_without_disturbances}
        \subcaption{}\label{fig:tracking_ellipse_with_disturbances}
        \subcaption{}\label{fig:planning_with_obstacle_experiment}
    \fi
    \caption{
        Trajectory plots for the tasks of:
        \protect(\subref{fig:tracking_circle_traj}) tracking a circular trajectory;
        \protect(\subref{fig:tracking_eight_traj}) tracking an eight-shaped trajectory;
        \protect(\subref{fig:plannning_without_obstacles}) planning a trajectory to a target;
        \protect(\subref{fig:planning_with_obstacles}) planning a trajectory to a target in the presence of obstacles;
        \protect(\subref{fig:tracking_ellipse_without_disturbances}) tracking an ellipsoidal trajectory with the robot;
        \protect(\subref{fig:tracking_ellipse_with_disturbances}) tracking an ellipsoidal trajectory with the robot in the presence of disturbances;
        \protect(\subref{fig:planning_with_obstacle_experiment}) planning and tracking a trajectory to a target with the robot in the presence of an obstacle.
    }
    \label{fig:trajectory_plots}
\end{figure*}

\subsection{Numerical Experiments --- Tracking Nominal Trajectory}

\paraDraft{Detail first experiment}
This experiment investigates the feasibility of the proposed~\acrshort*{mpcc} formulation for tracking in an~\acrshort*{mpc} loop.
We compare the~\acrshort*{mpcc} against the~\acrshort*{miqp}~\cite{Hogan2020}, for the task of tracking two different pre-specified nominal trajectories, a circular and an eight-shaped one. 
We generate pairs of nominal states and actions, which is a requirement from the~\acrshort*{miqp} (the~\acrshort*{mpcc} only requires nominal states), for a pushing motion with a sticking contact.
Both nominal trajectories have a total duration of~$10\,\si{\second}$ with sampling times of~$\Delta t=1/25\,\si{\second}$.
The sliding object is rectangular with dimensions~$7\times 12\,\si{\centi\meter}$ and the pusher has a radius of~$1\,\si{\centi\meter}$.

\begin{figure*}[t]
    \vspace{2mm}
    \centering
    \ifdefined\draftTikz
        \tikzsetnextfilename{fig_results_plots}
        \input{Tikz/fig_results_plots.tex}
        \vspace{-6mm}
    \else
        \includegraphics{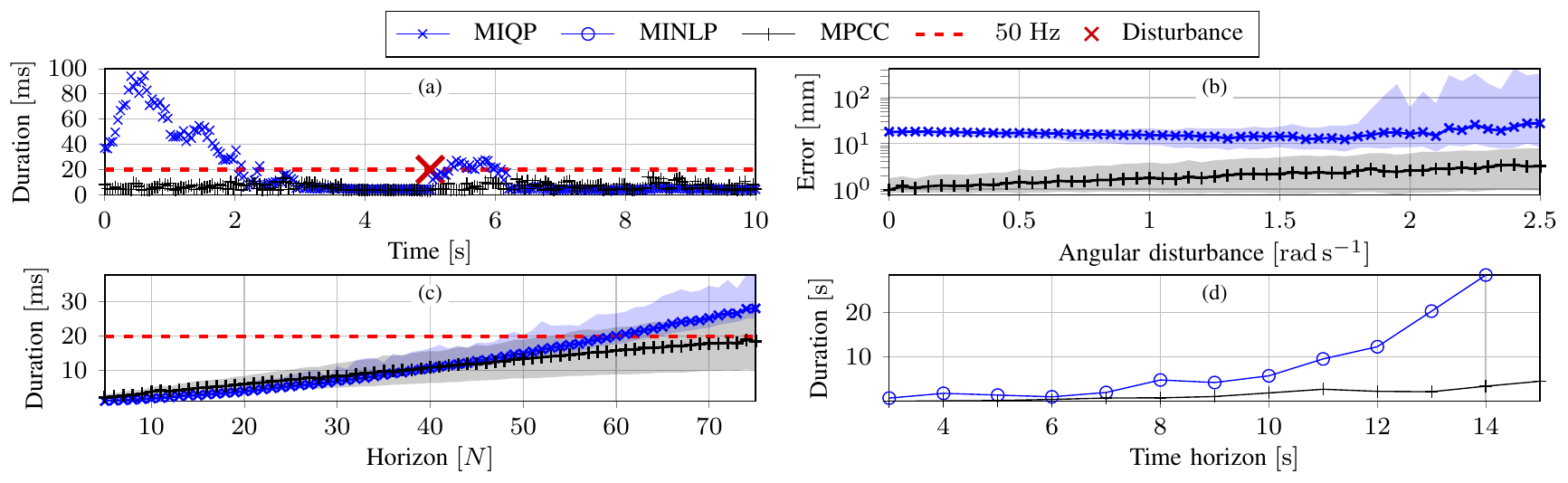}
    \fi
    \ifx\draftTikz\undefined
        \captionsetup[subfigure]{labelformat=empty}
        \subcaption{}\label{fig:tracking_circle_comp_time}
        \subcaption{}\label{fig:tracking_circle_disturbance_error}
        \subcaption{}\label{fig:tracking_circle_time_horizon_comp_time}
        \subcaption{}\label{fig:planning_comp_time}
    \fi
    \caption{
        Result plots of the:
        \protect(\subref{fig:tracking_circle_comp_time}) solving time while tracking a circular trajectory;
        \protect(\subref{fig:tracking_circle_disturbance_error}) trajectory tracking error when subject to angular disturbances, where we use a log scale for the vertical axis;
        \protect(\subref{fig:tracking_circle_time_horizon_comp_time}) computation time of solving the~\acrshort*{mpc} problem for different horizons;
        \protect(\subref{fig:planning_comp_time}) average solving time, over five runs, for planning the path to a given goal, as shown in~\cref{fig:plannning_without_obstacles}---where for~$T=15\,\si{s}$, the computation time of~\acrshort*{minlp} takes several minutes.
        Note that \protect(\subref{fig:tracking_circle_disturbance_error}-\subref{fig:tracking_circle_time_horizon_comp_time}) result from tracking the circular trajectory in~\cref{fig:tracking_circle_traj} ten times, corresponding to running the~\acrshort*{mpc} loop 2500 times per value in the horizontal axis, where the line corresponds to the median and the shaded area to the~\nth{10} and the~\nth{90} percentiles.
    }
    \label{fig:results_plots}
\end{figure*}

\paraDraft{Circle tracking}
Fig.~\ref{fig:tracking_circle_traj} shows the tracking of the circular trajectory.
We offset the initial state to~$x_0=[-0.03\,\si{\meter}, 0.03\,\si{\meter},30\,\si{\degree},0]^\top$ and at the~\nth{5} second introduce a disturbance of~$\Delta x=[0.03\,\si{\meter},-0.03\,\si{\meter},30\,\si{\degree},0]^\top$.
Both~\acrshort*{mpcc} and~\acrshort*{miqp} formulations are able to track the nominal trajectory, however the~\acrshort*{mpcc} recovers quicker after both the initial state offset and the disturbance.
Fig.~\ref{fig:tracking_circle_comp_time} shows the computation times for this experiment.
In the absence of disturbances both methods are able to compute solutions faster than $20\,\si{\milli\second}$ ($>50\,\si{\hertz}$).
However, when large deviations from the nominal trajectory occur the~\acrshort*{miqp} computation time increases significantly---rendering this method inappropriate for online use with large disturbances.
Additionally, we noticed that~\acrshort*{miqp} easily becomes unstable under larger disturbances unlike the~\acrshort*{mpcc}.

\paraDraft{Statistical comparisons}
Furthermore, we compare the~\acrshort*{mpcc} and the~\acrshort*{miqp} controllers for multiple runs on the circular trajectory when subject to angular disturbances, but without initial state offset.
In this experiment, we add a disturbance to the dynamics~\eqref{eq:system_dynamics} as~$\dot{x}=f+\epsilon$, where~$\epsilon=\begin{bmatrix}0,0,\epsilon_\theta,0\end{bmatrix}^\top$ and we draw~$\epsilon_\theta\sim\mathcal{U}(-\omega_M,\omega_M)$ from an uniform distribution.
\cref{fig:tracking_circle_disturbance_error} shows the evolution of the position trajectory error, computed as the distance between each point of the actual and the nominal trajectories, for increasing~$\omega_M$, where for each value of~$\omega_M$ we ran both controllers for tracking ten full circles, equating to a total of 2500~\acrshort*{mpc} loops.
\cref{fig:tracking_circle_disturbance_error} shows that the~\acrshort*{mpcc} results in a significant lower tracking error.
For the same experiment, but now with a constant~$\omega_M=1.5\;\si{\radian\per\second}$, we vary the horizon~$N$, and compute the~\acrshort*{mpc} computation time.
\cref{fig:tracking_circle_time_horizon_comp_time} shows that the~\acrshort*{mpcc} scales better.

\begin{figure*}[t]
    \centering
    \begin{subfigure}{0.155\textwidth}
        \includegraphics[width=\textwidth]{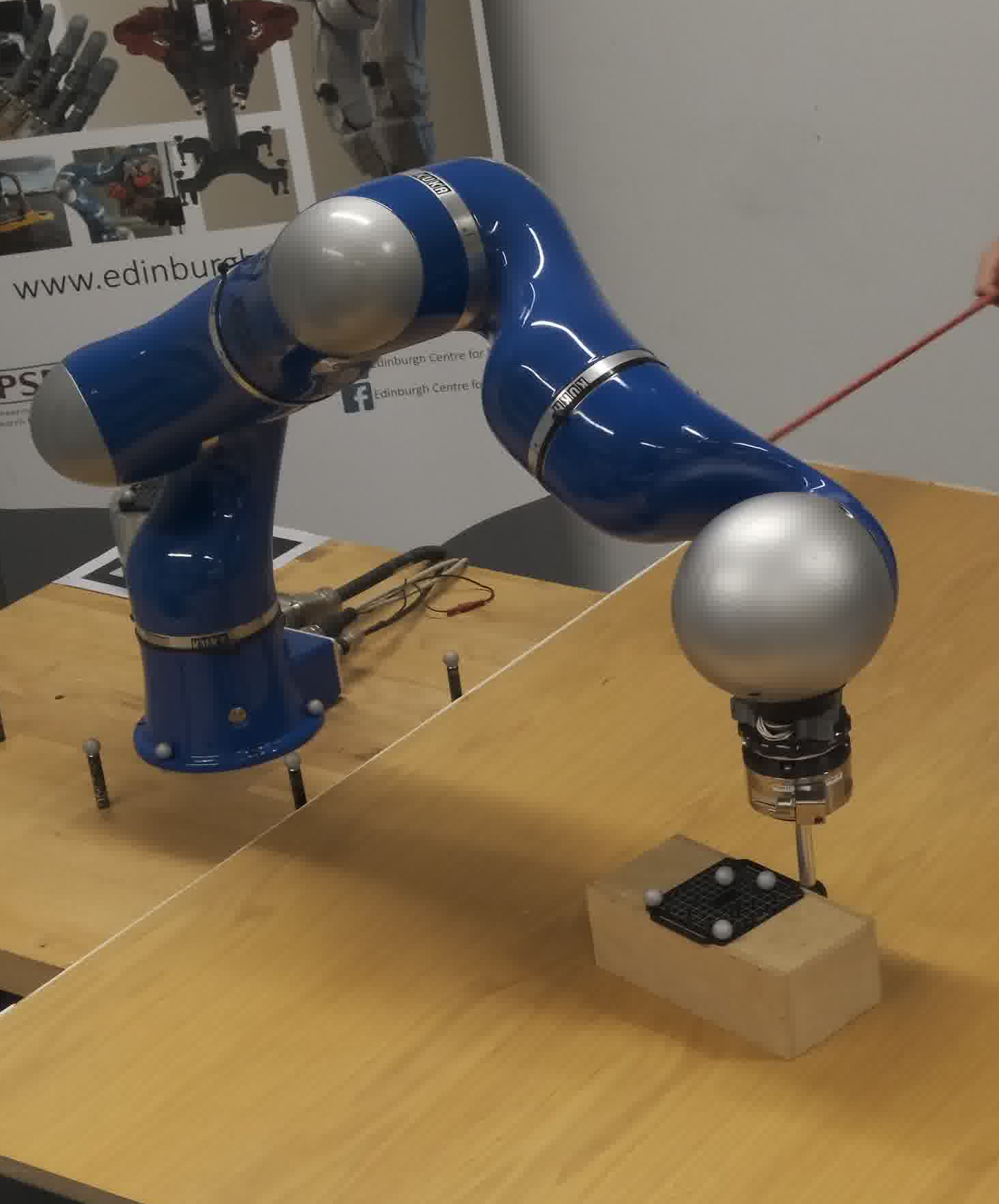}
        \vspace{-5mm}
        \caption{}
        \label{fig:exp_keyframe_a}
    \end{subfigure}
    \begin{subfigure}{0.155\textwidth}
        \includegraphics[width=\textwidth]{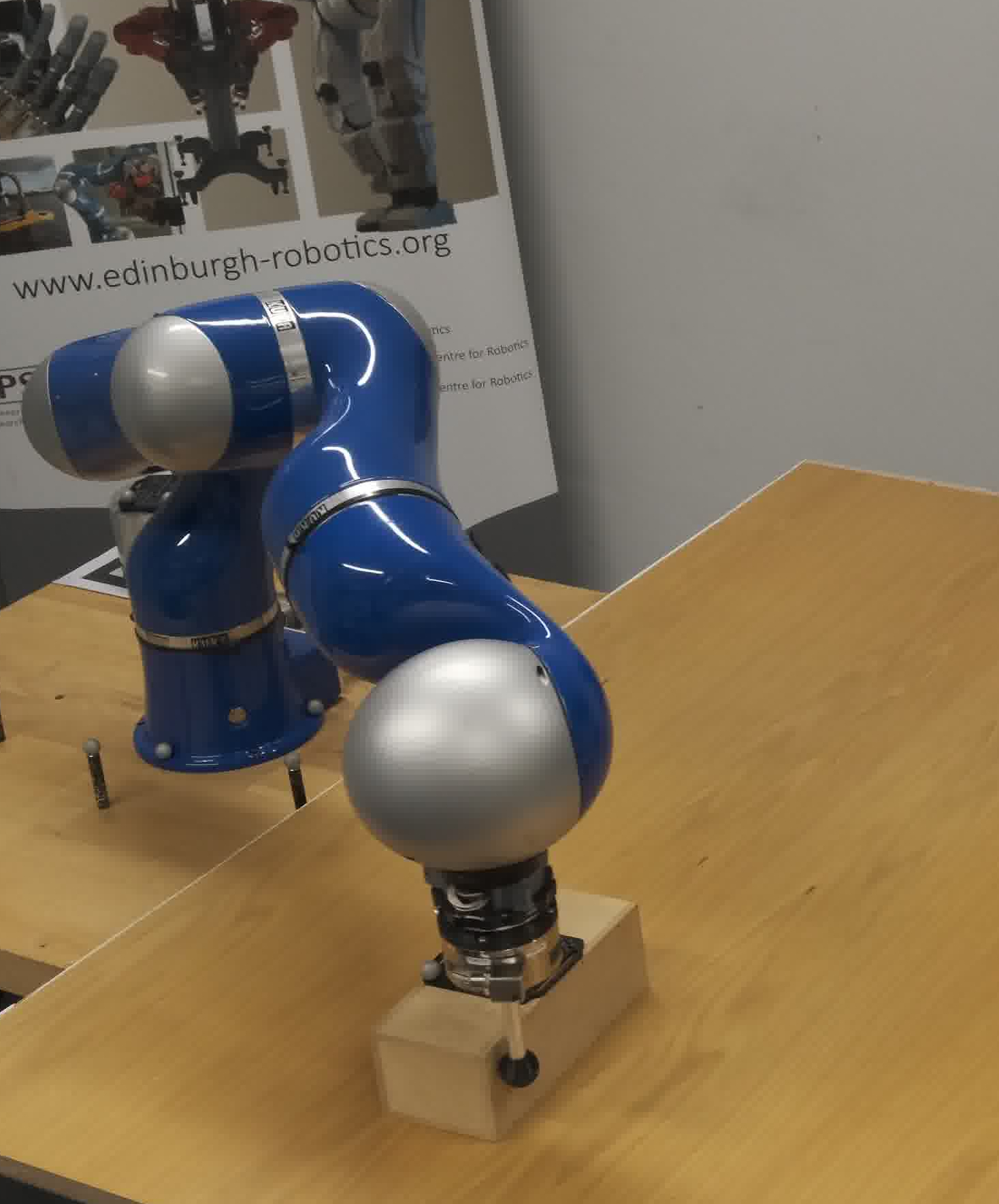}
        \vspace{-5mm}
        \caption{}
        \label{fig:exp_keyframe_b}
    \end{subfigure}
    \begin{subfigure}{0.155\textwidth}
        \includegraphics[width=\textwidth]{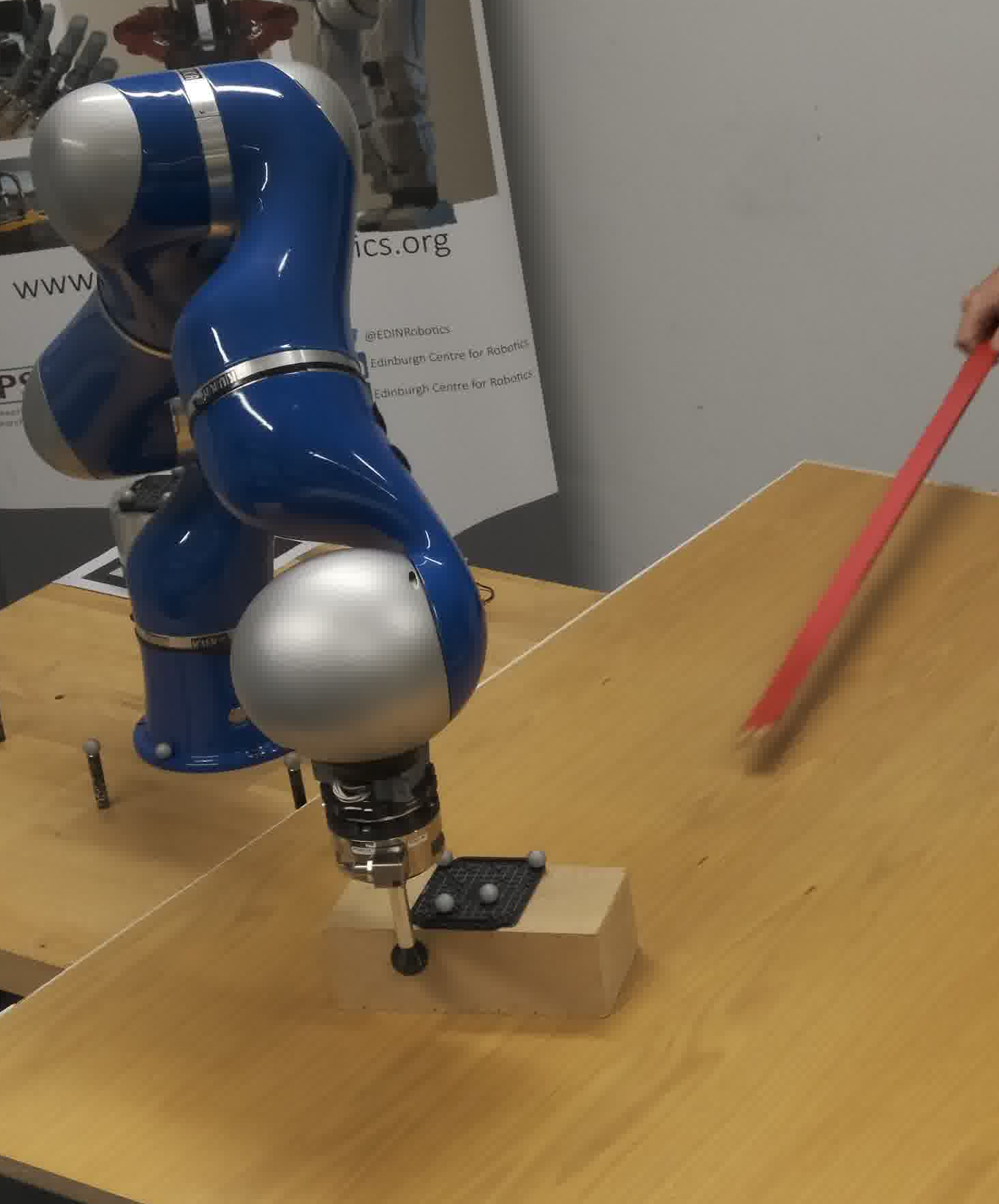}
        \vspace{-5mm}
        \caption{}
        \label{fig:exp_keyframe_c}
    \end{subfigure}
    \begin{subfigure}{0.155\textwidth}
        \includegraphics[width=\textwidth]{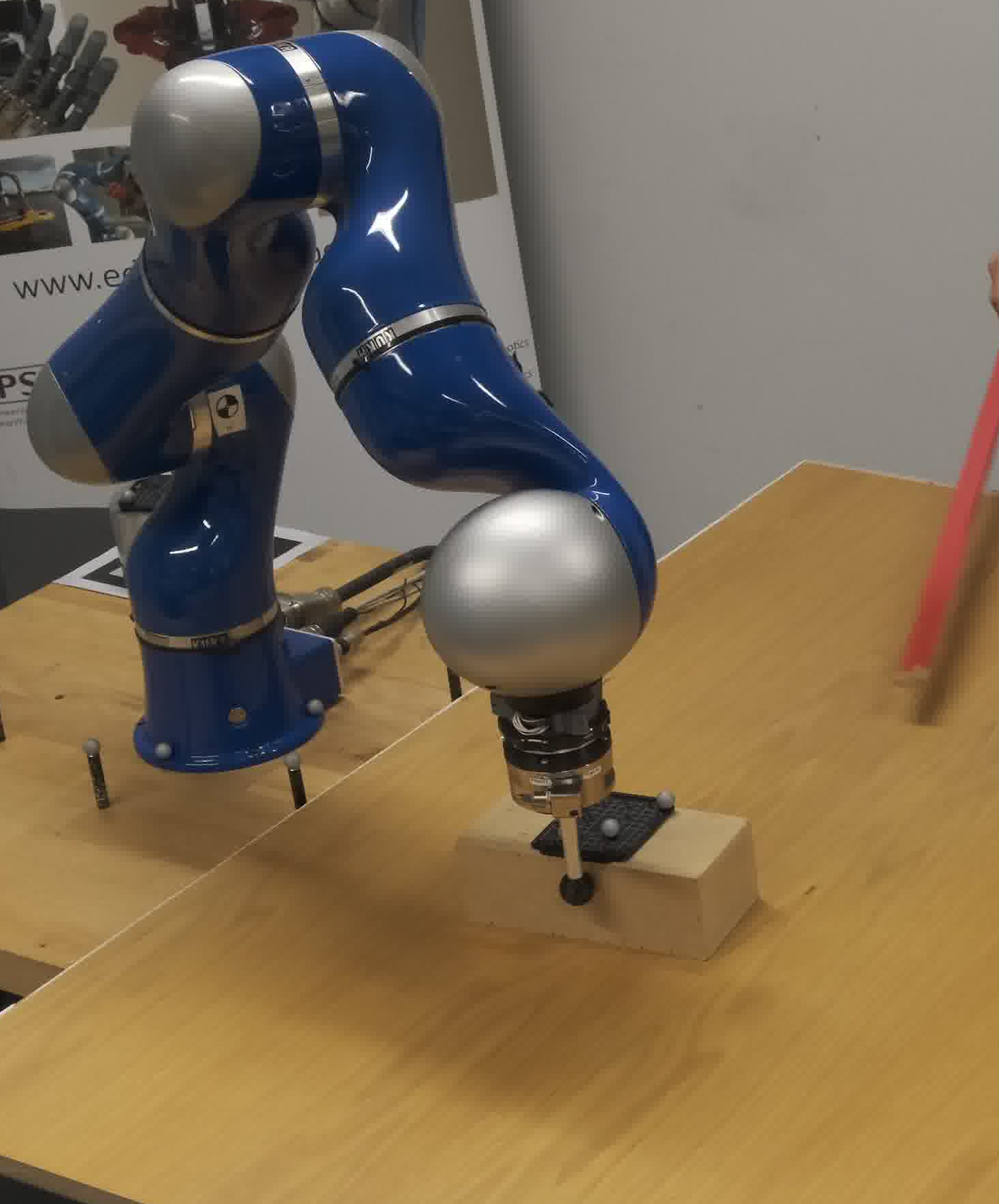}
        \vspace{-5mm}
        \caption{}
        \label{fig:exp_keyframe_d}
    \end{subfigure}
    \begin{subfigure}{0.155\textwidth}
        \includegraphics[width=\textwidth]{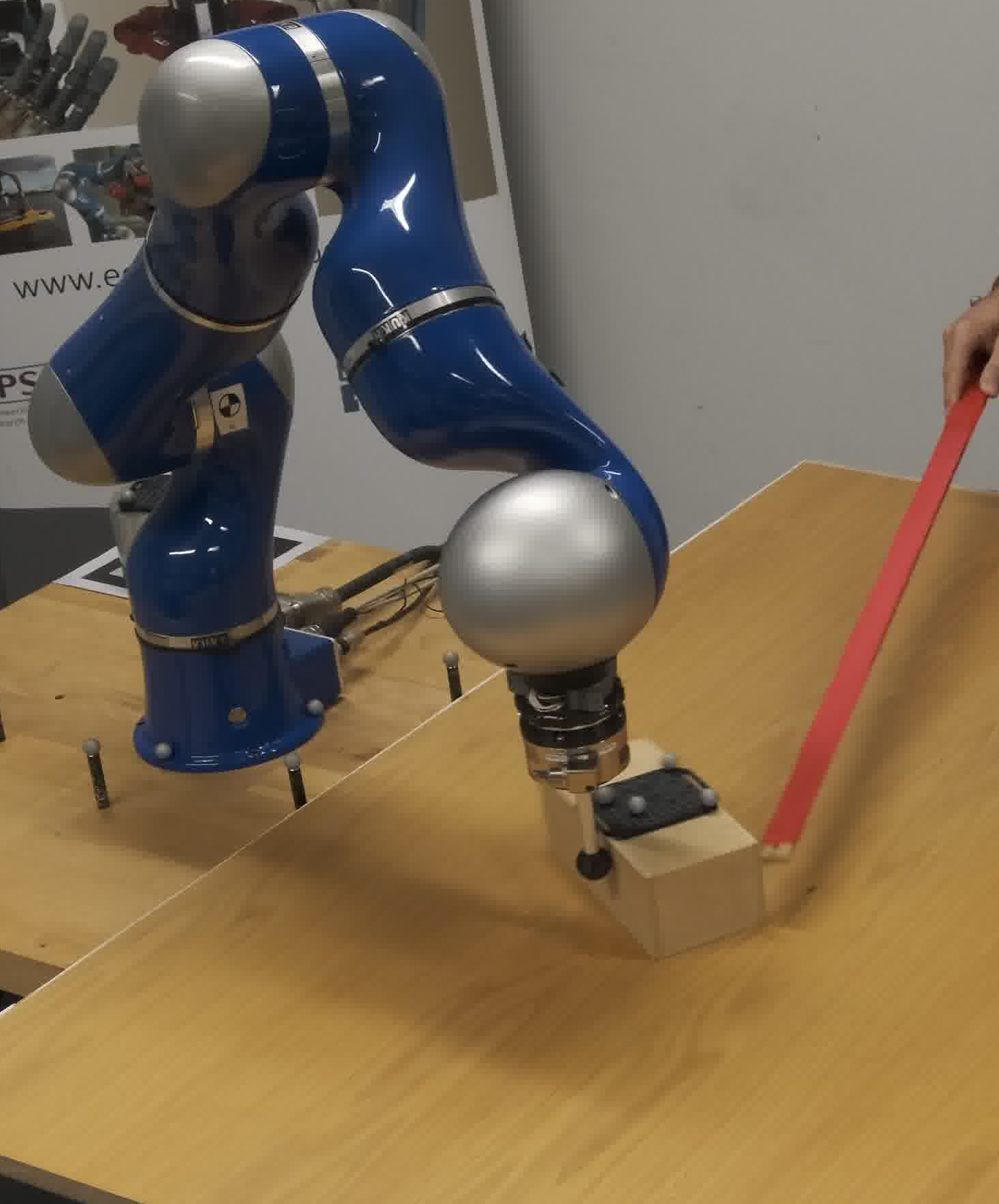}
        \vspace{-5mm}
        \caption{}
        \label{fig:exp_keyframe_e}
    \end{subfigure}
    \begin{subfigure}{0.155\textwidth}
        \includegraphics[width=\textwidth]{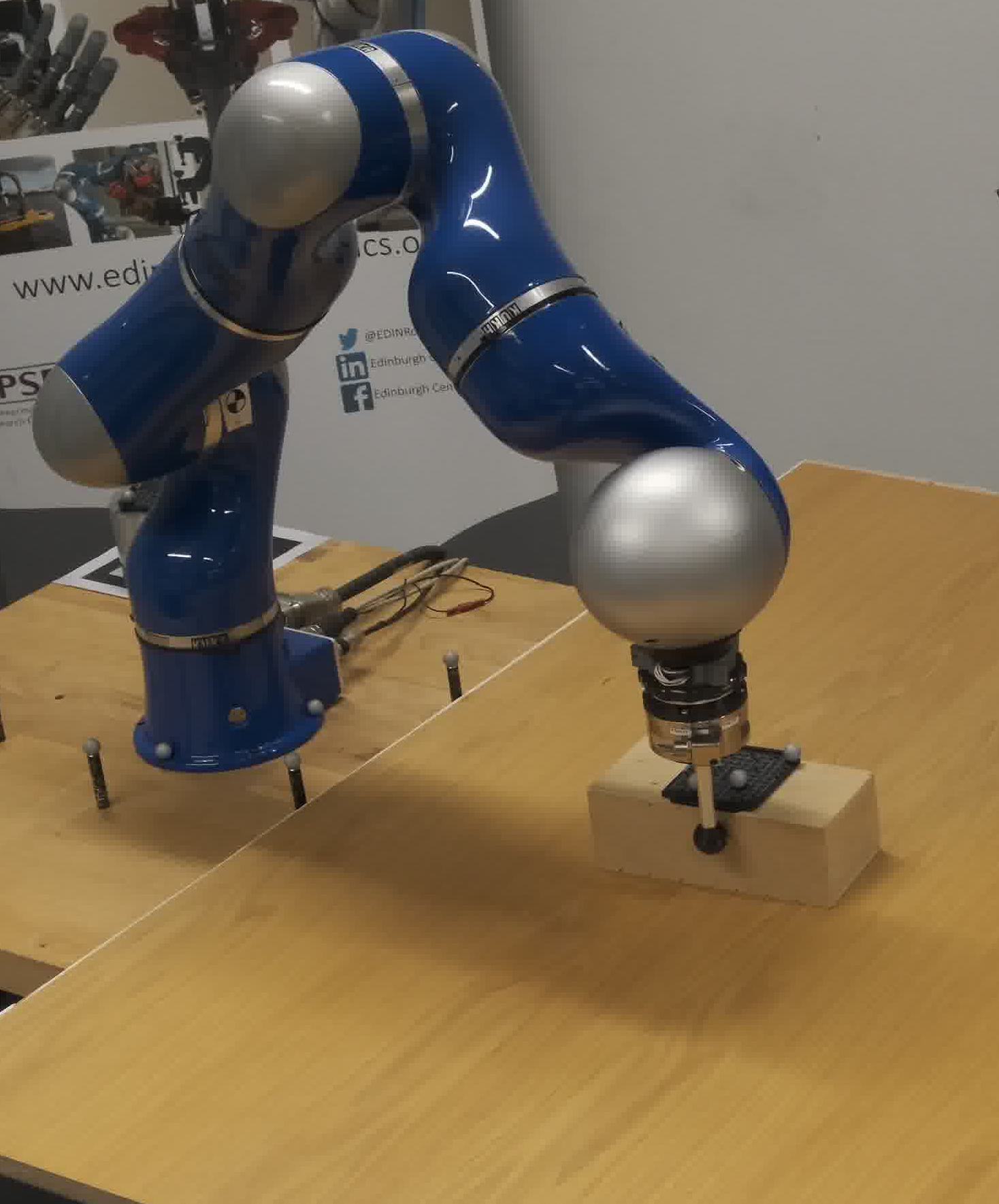}
        \vspace{-5mm}
        \caption{}
        \label{fig:exp_keyframe_f}
    \end{subfigure}
    \vspace{2mm}
    \caption{Keyframes of the KUKA LWR robot tracking an ellipsoidal trajectory, where \protect(\subref{fig:exp_keyframe_a}-\subref{fig:exp_keyframe_c}) show the sliding of the object along an aggressive curvature segment of the trajectory, and \protect(\subref{fig:exp_keyframe_d}-\subref{fig:exp_keyframe_f}) exhibit the \acrshort*{mpc}  recovering from a disturbance.}
    \label{fig:experimental_keyframe_results}
\end{figure*}

\paraDraft{Discuss eight tracking}
Fig.~\ref{fig:tracking_eight_traj} shows the tracking of the eight-shaped trajectory, showing that the~\acrshort*{mpcc} achieves qualitatively better tracking than the~\acrshort*{miqp}.
Note that the~\acrshort*{miqp} linearises the dynamics along the nominal states and actions, which makes its tracking quality highly depend on the accuracy of the nominal state and action pairs. 
However, obtaining accurate nominal actions in scenarios with aggressive curvature, like the one shown in Fig.~\ref{fig:tracking_eight_traj}, is often unachievable.
In contrast to~\acrshort*{miqp}, the~\acrshort*{mpcc} dispenses with both the linearization and the nominal actions, which might explain its improved tracking.

\subsection{Numerical Experiments --- Planning}

\paraDraft{Experiment goal}
In this experiment, we test the capabilities of the~\acrshort*{mpcc} formulation to generate trajectories by exploiting sticking and sliding contact modes.
These trajectories drive the sliding object from an initial state to a desired target state---even in the presence of obstacles.
We compare the~\acrshort*{mpcc} formulation against the~\acrshort*{minlp}, see~\eqref{eq:minlp}, both in terms of convergence and computation time. 
The planning tasks have as initial state~$x_0=[0, 0, 0, 0]^\top$ and as target state~$x^\ast_N=[0.3\,\si{\meter}, 0.4\,\si{\meter}, 270\,\si{\degree}, 0]^\top$.
For the scenario with the obstacles each obstacle has a radius of~$5\,\si{\milli\meter}$.
We encode the obstacles in~\eqref{eq:mpcc} as inequality constraints on the distance between the obstacle and the center of the sliding object.

\paraDraft{Experiment discussion}
Fig.~\ref{fig:plannning_without_obstacles} shows an example of a trajectory generated for a~\acrshort*{th} of~$T=3\,\si{\second}$.
Additionally, we generated trajectories for the same task but with varying~\acrshort*{th}, from~$3\,\si{\second}$ to~$15\,\si{\second}$.
Fig.~\ref{fig:planning_comp_time} shows the respective computation times, showing that the~\acrshort*{mpcc} takes significantly less time than the~\acrshort*{minlp} to converge and scales well with respect to the~\acrshort*{th}.
Furthermore, for the scenario including obstacles, the~\acrshort*{minlp} always failed to converge to a solution, while the~\acrshort*{mpcc} was able to produce plans, like the one shown in Fig.~\ref{fig:planning_with_obstacles}.
Note, however, that the~\acrshort*{mpcc} optimization was able to produce planning solutions at the cost of non zero values for the complementarity slack variables, unlike the case when using~\acrshort*{mpcc} for tracking with~\acrshort*{mpc}, which always gave solutions with zero slack.

\subsection{Robot Experiment --- Tracking Nominal Trajectory}\label{subsec:robot_exp_tracking}

\paraDraft{Introduce experiment}
In this experiment, we used the KUKA LWR robot to assess our~\acrshort*{mpc} implementation of the~\acrshort*{mpcc} formulation for the task of tracking an ellipsoidal trajectory.
The radius of the pusher is~$16\,\si{\milli\meter}$ and the size of the box is~$9\times 20\,\si{\centi\meter}$.

\paraDraft{Report results}
Fig.~\ref{fig:tracking_ellipse_without_disturbances} and~\ref{fig:tracking_ellipse_with_disturbances} depict two tracking experiments, one without and one with disturbances, respectively.
Fig.~\ref{fig:experimental_keyframe_results} shows a sequence of images of the robot tracking the ellipsoidal trajectory, where~\ref{fig:exp_keyframe_a}--\ref{fig:exp_keyframe_c} correspond to the moment when the robot turns the object around one of the corners of the ellipse using a sliding motion, and~\ref{fig:exp_keyframe_d}--\ref{fig:exp_keyframe_f} correspond to the moment when the robot recovers from a disturbance.
In the accompanying video, we also show the~\acrshort*{mpc} modestly handling dynamic obstacles.
Due to its short sighed~\acrshort*{th}---only~$1\,\si{\second}$---it only reacts when the obstacle is imminent.

\subsection{Robot Experiment --- Planning}

\paraDraft{Describe experiment}
In the final experiment, we demonstrate both planning a trajectory with obstacles and tracking that trajectory using the proposed~\acrshort*{mpcc} formulation.
The planning~\acrshort*{th} was~$T=22\,\si{\second}$.
Fig.~\ref{fig:planning_with_obstacle_experiment} shows both the planning and the tracking trajectories for both the slider and the pusher.
This experiment exemplifies the flexibility of our~\acrshort*{mpcc} formulation, which is able to both plan and control the non-holonomic pusher-slider system---just by adjusting the~\acrshort*{th} of the problem.

%% file: Tikz/fig_trajectory_plots.tex
\def\xLSim{0.07}
\def\yLSim{0.12}
\def\rpSim{0.01}
\def\idxStart{0}
\begin{tikzpicture}
    \input{Tikz/tikz.tex}
    \pgfplotstableread[col sep=comma, header=true]{data/tracking_circle_cc_state.csv}{\tableTrackingCircleCCState}
    \pgfplotstableread[col sep=comma, header=true]{data/tracking_circle_mi_state.csv}{\tableTrackingCircleMIState}
    \pgfplotstablegetelem{\idxStart}{x_opt}\of\tableTrackingCircleCCState
    \pgfmathsetmacro{\xStart}{\pgfplotsretval}
    \pgfplotstablegetelem{\idxStart}{y_opt}\of\tableTrackingCircleCCState
    \pgfmathsetmacro{\yStart}{\pgfplotsretval}
    \pgfplotstablegetelem{\idxStart}{theta_opt}\of\tableTrackingCircleCCState
    \pgfmathsetmacro{\thetaStart}{\pgfplotsretval}
    \pgfplotstablegetelem{\idxStart}{x_pusher}\of\tableTrackingCircleCCState
    \pgfmathsetmacro{\xpStart}{\pgfplotsretval}
    \pgfplotstablegetelem{\idxStart}{y_pusher}\of\tableTrackingCircleCCState
    \pgfmathsetmacro{\ypStart}{\pgfplotsretval}
    \begin{axis}[
        name=circle,
        xlabel={$\mathrm{x}\;[\si{\meter}]$},
        ylabel={$\mathrm{y}\;[\si{\meter}]$},
        enlarge x limits=0.05,
        enlarge y limits=0.05,
        axis equal image,
        height=6.0cm,
        legend style={column sep=2mm},
        legend to name={commonLegendTraj},
        legend columns = 3,
        transpose legend,
    ]
        \addlegendimage{black!50!green,->,very thick}
        \addlegendentry{Slider initial config.}
        \addlegendimage{black!50!green,only marks,mark=*}
        \addlegendentry{Pusher initial pos.}
        \addlegendimage{black!20!red,->,very thick}
        \addlegendentry{Slider goal config.}
        \addlegendimage{black!20!red,only marks,mark=*}
        \addlegendentry{Pusher goal config.}
        \addlegendimage{black,->,very thick}
        \addlegendentry{Slider config.}
        \addlegendimage{black,only marks,mark=*}
        \addlegendentry{Pusher pos.}
        \addlegendimage{dashdotted, thick, black!20!red}
        \addlegendentry{Slider nominal traj.}
        \addlegendimage{black!20!red,thick}
        \addlegendentry{Slider planned traj.}
        \addlegendimage{thick, black!20!red, dashed}
        \addlegendentry{Pusher planned traj.}
        \addlegendimage{black}
        \addlegendentry{Slider MPCC traj.}
        \addlegendimage{dashed, black}
        \addlegendentry{Pusher MPCC traj.}
        \addlegendimage{densely dashdotdotted, blue}
        \addlegendentry{Slider MIQP traj.}
        \addlegendimage{blue!80!black,only marks,mark=*,mark size=4.0pt}
        \addlegendentry{Obstacle}
        \addlegendimage{mark=x,only marks,black!20!red,mark size=3.0pt,thick}
        \addlegendentry{Disturbance}
        \coordinate (coorStart) at (axis cs:\xStart,\yStart);
        \pic {rectStart={coorStart,deg(\thetaStart),\xLSim,\yLSim}};
        \draw[fill=black!50!green,draw=black!50!green] \pgfextra{
            \pgfpathellipse{\pgfplotspointaxisxy{\xpStart}{\ypStart}}
            {\pgfplotspointaxisdirectionxy{\rpSim}{0}}
            {\pgfplotspointaxisdirectionxy{0}{\rpSim}}
        };
        \pgfplotsextra{
            \foreach \i in {15,140}{
                \pgfplotstablegetelem{\i}{x_opt}\of\tableTrackingCircleCCState
                \pgfmathsetmacro{\xCoor}{\pgfplotsretval}
                \pgfplotstablegetelem{\i}{y_opt}\of\tableTrackingCircleCCState
                \pgfmathsetmacro{\yCoor}{\pgfplotsretval}
                \pgfplotstablegetelem{\i}{theta_opt}\of\tableTrackingCircleCCState
                \pgfmathsetmacro{\theta}{\pgfplotsretval};
                \pgfplotstablegetelem{\i}{x_pusher}\of\tableTrackingCircleCCState
                \pgfmathsetmacro{\xp}{\pgfplotsretval};
                \pgfplotstablegetelem{\i}{y_pusher}\of\tableTrackingCircleCCState
                \pgfmathsetmacro{\yp}{\pgfplotsretval};
                \coordinate (coor) at (axis cs:\xCoor,\yCoor);
                \coordinate (coorp) at (axis cs:\xp,\yp);
                \pic {rectObj={coor,deg(\theta),\xLSim,\yLSim}};
                \draw[fill=black,draw=black] \pgfextra{
                    \pgfpathellipse{\pgfplotspointaxisxy{\xp}{\yp}}
                    {\pgfplotspointaxisdirectionxy{\rpSim}{0}}
                    {\pgfplotspointaxisdirectionxy{0}{\rpSim}}
                };
            }
        }
        \addplot[
            dashdotted,
            thick,
            black!20!red,
        ] table [
            col sep=comma,
            x={x_nom},
            y={y_nom},
        ] {\tableTrackingCircleCCState};
        \addplot[
            black,
        ] table [
            col sep=comma,
            x={x_opt},
            y={y_opt},
        ] {\tableTrackingCircleCCState};
        \addplot[
            densely dashdotdotted,
            blue,
        ] table [
            col sep=comma,
            x={x_opt},
            y={y_opt},
        ] {\tableTrackingCircleMIState};
        \pgfplotstablegetelem{125}{x_opt}\of\tableTrackingCircleCCState
        \pgfmathsetmacro{\xCoor}{\pgfplotsretval}
        \pgfplotstablegetelem{125}{y_opt}\of\tableTrackingCircleCCState
        \pgfmathsetmacro{\yCoor}{\pgfplotsretval}
        \coordinate (coor) at (axis cs:\xCoor,\yCoor);
        \draw[color=black!20!red, mark size=4.0pt, thick] plot[mark=x] coordinates{(coor)} node {};
        \pgfplotstablegetelem{125}{x_opt}\of\tableTrackingCircleMIState
        \pgfmathsetmacro{\xCoorb}{\pgfplotsretval}
        \pgfplotstablegetelem{125}{y_opt}\of\tableTrackingCircleMIState
        \pgfmathsetmacro{\yCoorb}{\pgfplotsretval}
        \coordinate (coorb) at (axis cs:\xCoorb,\yCoorb);
        \draw[color=black!20!red, mark size=4.0pt, thick] plot[mark=x] coordinates{(coorb)} node {};
    \end{axis}
    \pgfplotstableread[col sep=comma, header=true]{data/tracking_eight_cc_state.csv}{\tableTrackingEightCCState}
    \pgfplotstableread[col sep=comma, header=true]{data/tracking_eight_mi_state.csv}{\tableTrackingEightMIState}
    \pgfplotstablegetelem{\idxStart}{x_opt}\of\tableTrackingEightCCState
    \pgfmathsetmacro{\xStart}{\pgfplotsretval}
    \pgfplotstablegetelem{\idxStart}{y_opt}\of\tableTrackingEightCCState
    \pgfmathsetmacro{\yStart}{\pgfplotsretval}
    \pgfplotstablegetelem{\idxStart}{theta_opt}\of\tableTrackingEightCCState
    \pgfmathsetmacro{\thetaStart}{\pgfplotsretval}
    \pgfplotstablegetelem{\idxStart}{x_pusher}\of\tableTrackingEightCCState
    \pgfmathsetmacro{\xpStart}{\pgfplotsretval}
    \pgfplotstablegetelem{\idxStart}{y_pusher}\of\tableTrackingEightCCState
    \pgfmathsetmacro{\ypStart}{\pgfplotsretval}
    \begin{axis}[
        name=eight,
        at={($(circle.east)+(0.9cm,0)$)},
        anchor=west,
        height=6.0cm,
        xlabel={$\mathrm{x}\;[\si{\meter}]$},
        xtick={-0.1,0,0.1},
        enlarge x limits=0.05,
        enlarge y limits=0.05,
        axis equal image,
    ]
        \pgfmathsetmacro{\thetaStartRot}{deg(\thetaStart)+90}
        \pgfmathsetmacro{\xpStartRot}{-\ypStart}
        \pgfmathsetmacro{\ypStartRot}{\xpStart}
        \coordinate (coorStart) at (axis cs:\xStart,\yStart);
        \pic {rectStart={coorStart,\thetaStartRot,\xLSim,\yLSim}};
        \draw[fill=black!50!green,draw=black!50!green] \pgfextra{
            \pgfpathellipse{\pgfplotspointaxisxy{\xpStartRot}{\ypStartRot}}
            {\pgfplotspointaxisdirectionxy{\rpSim}{0}}
            {\pgfplotspointaxisdirectionxy{0}{\rpSim}}
        };
        \pgfplotsextra{
            \foreach \i in {45,210}{%
                \pgfplotstablegetelem{\i}{x_opt}\of\tableTrackingEightCCState
                \pgfmathsetmacro{\xCoor}{\pgfplotsretval}
                \pgfplotstablegetelem{\i}{y_opt}\of\tableTrackingEightCCState
                \pgfmathsetmacro{\yCoor}{\pgfplotsretval}
                \pgfplotstablegetelem{\i}{theta_opt}\of\tableTrackingEightCCState
                \pgfmathsetmacro{\theta}{\pgfplotsretval};
                \pgfplotstablegetelem{\i}{x_pusher}\of\tableTrackingEightCCState
                \pgfmathsetmacro{\xp}{\pgfplotsretval};
                \pgfplotstablegetelem{\i}{y_pusher}\of\tableTrackingEightCCState
                \pgfmathsetmacro{\yp}{\pgfplotsretval};
                \pgfmathsetmacro{\thetaR}{deg(\theta)+90}
                \pgfmathsetmacro{\xCoorR}{-\yCoor}
                \pgfmathsetmacro{\yCoorR}{\xCoor}
                \pgfmathsetmacro{\xpR}{-\yp}
                \pgfmathsetmacro{\ypR}{\xp}
                \coordinate (coor) at (axis cs:\xCoorR,\yCoorR);
                \coordinate (coorp) at (axis cs:\xpR,\ypR);
                \pic {rectObj={coor,\thetaR,\xLSim,\yLSim}};
                \draw[fill=black,draw=black] \pgfextra{
                    \pgfpathellipse{\pgfplotspointaxisxy{\xpR}{\ypR}}
                    {\pgfplotspointaxisdirectionxy{0.008}{0}}
                    {\pgfplotspointaxisdirectionxy{0}{0.008}}
                };
            }
        }
        \addplot[
            black!20!red,
            dashdotted,
            thick,
        ] table [
            col sep=comma,
            x expr={-\thisrow{y_nom}},
            y={x_nom},
        ] {\tableTrackingEightCCState};
        \addplot[
            black,
        ] table [
            col sep=comma,
            x expr={-\thisrow{y_opt}},
            y={x_opt},
        ] {\tableTrackingEightCCState};
        \addplot[
            densely dashdotdotted,
            blue,
        ] table [
            col sep=comma,
            x expr={-\thisrow{y_opt}},
            y={x_opt},
        ] {\tableTrackingEightMIState};
    \end{axis}
    \def\idxMiddle{35}
    \pgfplotstableread[col sep=comma, header=true]{data/planning_without_obstacles_state.csv}{\table}
    \pgfplotstablegetrowsof{\table} 
    \pgfmathsetmacro{\Nrows}{\pgfplotsretval-1}
    \pgfplotstablegetelem{\idxStart}{x_slider}\of{\table}
    \pgfmathsetmacro{\xStart}{\pgfplotsretval}
    \pgfplotstablegetelem{\idxStart}{y_slider}\of{\table}
    \pgfmathsetmacro{\yStart}{\pgfplotsretval}
    \pgfplotstablegetelem{\idxStart}{theta_slider}\of{\table}
    \pgfmathsetmacro{\thetaStart}{\pgfplotsretval}
    \pgfplotstablegetelem{\idxStart}{x_pusher}\of{\table}
    \pgfmathsetmacro{\xpStart}{\pgfplotsretval}
    \pgfplotstablegetelem{\idxStart}{y_pusher}\of{\table}
    \pgfmathsetmacro{\ypStart}{\pgfplotsretval}
    \pgfplotstablegetelem{\idxMiddle}{x_slider}\of{\table}
    \pgfmathsetmacro{\xMiddle}{\pgfplotsretval}
    \pgfplotstablegetelem{\idxMiddle}{y_slider}\of{\table}
    \pgfmathsetmacro{\yMiddle}{\pgfplotsretval}
    \pgfplotstablegetelem{\idxMiddle}{theta_slider}\of{\table}
    \pgfmathsetmacro{\thetaMiddle}{\pgfplotsretval}
    \pgfplotstablegetelem{\idxMiddle}{x_pusher}\of{\table}
    \pgfmathsetmacro{\xpMiddle}{\pgfplotsretval}
    \pgfplotstablegetelem{\idxMiddle}{y_pusher}\of{\table}
    \pgfmathsetmacro{\ypMiddle}{\pgfplotsretval}
    \pgfplotstablegetelem{\Nrows}{x_slider}\of{\table}
    \pgfmathsetmacro{\xGoal}{\pgfplotsretval}
    \pgfplotstablegetelem{\Nrows}{y_slider}\of{\table}
    \pgfmathsetmacro{\yGoal}{\pgfplotsretval}
    \pgfplotstablegetelem{\Nrows}{theta_slider}\of{\table}
    \pgfmathsetmacro{\thetaEnd}{\pgfplotsretval}
    \pgfplotstablegetelem{\Nrows}{x_pusher}\of{\table}
    \pgfmathsetmacro{\xpGoal}{\pgfplotsretval}
    \pgfplotstablegetelem{\Nrows}{y_pusher}\of{\table}
    \pgfmathsetmacro{\ypGoal}{\pgfplotsretval}
    \begin{axis}[
        name=planningWithoutObs,
        at={($(circle.south west)+(0,-0.9cm)$)},
        anchor=north west,
        xlabel={$\mathrm{x}\;[\si{\meter}]$},
        ylabel={$\mathrm{y}\;[\si{\meter}]$},
        height=6.0cm,
        xmin=-0.07, xmax=0.43,
        axis equal image,
    ]
        \coordinate (coorStart) at (axis cs:\xStart,\yStart);
        \coordinate (coorMiddle) at (axis cs:\xMiddle,\yMiddle);
        \coordinate (coorEnd) at (axis cs:\xGoal,\yGoal);
        \pic {rectStart={coorStart,deg(\thetaStart),\xLSim,\yLSim}};
        \pic {rectObj={coorMiddle,deg(\thetaMiddle),\xLSim,\yLSim}};
        \pic {rectGoal={coorEnd,deg(\thetaEnd),\xLSim,\yLSim}};
        \draw[fill=black!50!green,draw=black!50!green] \pgfextra{
            \pgfpathellipse{\pgfplotspointaxisxy{\xpStart}{\ypStart}}
            {\pgfplotspointaxisdirectionxy{0.01}{0}}
            {\pgfplotspointaxisdirectionxy{0}{0.01}}
        };
        \draw[fill=black,draw=black] \pgfextra{
            \pgfpathellipse{\pgfplotspointaxisxy{\xpMiddle}{\ypMiddle}}
            {\pgfplotspointaxisdirectionxy{0.01}{0}}
            {\pgfplotspointaxisdirectionxy{0}{0.01}}
        };
        \draw[fill=black!20!red,draw=black!20!red] \pgfextra{
            \pgfpathellipse{\pgfplotspointaxisxy{\xpGoal}{\ypGoal}}
            {\pgfplotspointaxisdirectionxy{0.01}{0}}
            {\pgfplotspointaxisdirectionxy{0}{0.01}}
        };
        \addplot[
            black!20!red,
            thick,
        ] table [
            col sep=comma,
            x={x_slider},
            y={y_slider},
        ] {\table};
        \addplot[
            dashed,
            black!20!red,
            thick,
        ] table [
            col sep=comma,
            x={x_pusher},
            y={y_pusher},
        ] {\table};
    \end{axis}
    \def\idxMiddle{35}
    \pgfplotstableread[col sep=comma, header=true]{data/planning_with_obstacles_state.csv}{\table}
    \pgfplotstablegetrowsof{\table} 
    \pgfmathsetmacro{\Nrows}{\pgfplotsretval-1}
    \pgfplotstablegetelem{\idxStart}{x_slider}\of{\table}
    \pgfmathsetmacro{\xStart}{\pgfplotsretval}
    \pgfplotstablegetelem{\idxStart}{y_slider}\of{\table}
    \pgfmathsetmacro{\yStart}{\pgfplotsretval}
    \pgfplotstablegetelem{\idxStart}{theta_slider}\of{\table}
    \pgfmathsetmacro{\thetaStart}{\pgfplotsretval}
    \pgfplotstablegetelem{\idxStart}{x_pusher}\of{\table}
    \pgfmathsetmacro{\xpStart}{\pgfplotsretval}
    \pgfplotstablegetelem{\idxStart}{y_pusher}\of{\table}
    \pgfmathsetmacro{\ypStart}{\pgfplotsretval}
    \pgfplotstablegetelem{\idxMiddle}{x_slider}\of{\table}
    \pgfmathsetmacro{\xMiddle}{\pgfplotsretval}
    \pgfplotstablegetelem{\idxMiddle}{y_slider}\of{\table}
    \pgfmathsetmacro{\yMiddle}{\pgfplotsretval}
    \pgfplotstablegetelem{\idxMiddle}{theta_slider}\of{\table}
    \pgfmathsetmacro{\thetaMiddle}{\pgfplotsretval}
    \pgfplotstablegetelem{\idxMiddle}{x_pusher}\of{\table}
    \pgfmathsetmacro{\xpMiddle}{\pgfplotsretval}
    \pgfplotstablegetelem{\idxMiddle}{y_pusher}\of{\table}
    \pgfmathsetmacro{\ypMiddle}{\pgfplotsretval}
    \pgfplotstablegetelem{\Nrows}{x_slider}\of{\table}
    \pgfmathsetmacro{\xGoal}{\pgfplotsretval}
    \pgfplotstablegetelem{\Nrows}{y_slider}\of{\table}
    \pgfmathsetmacro{\yGoal}{\pgfplotsretval}
    \pgfplotstablegetelem{\Nrows}{theta_slider}\of{\table}
    \pgfmathsetmacro{\thetaEnd}{\pgfplotsretval}
    \pgfplotstablegetelem{\Nrows}{x_pusher}\of{\table}
    \pgfmathsetmacro{\xpGoal}{\pgfplotsretval}
    \pgfplotstablegetelem{\Nrows}{y_pusher}\of{\table}
    \pgfmathsetmacro{\ypGoal}{\pgfplotsretval}
    \begin{axis}[
        name=planningWithObs,
        at={($(eight.south east)+(0,-0.9cm)$)},
        anchor=north east,
        xlabel={$\mathrm{x}\;[\si{\meter}]$},
        height=6.0cm,
        xmin=-0.07, xmax=0.43,
        yticklabels={,,},
        axis equal image,
    ]
        \coordinate (coorStart) at (axis cs:\xStart,\yStart);
        \coordinate (coorMiddle) at (axis cs:\xMiddle,\yMiddle);
        \coordinate (coorEnd) at (axis cs:\xGoal,\yGoal);
        \pic {rectStart={coorStart,deg(\thetaStart),\xLSim,\yLSim}};
        \pic {rectObj={coorMiddle,deg(\thetaMiddle),\xLSim,\yLSim}};
        \pic {rectGoal={coorEnd,deg(\thetaEnd),\xLSim,\yLSim}};
        \draw[fill=black!50!green,draw=black!50!green] \pgfextra{
            \pgfpathellipse{\pgfplotspointaxisxy{\xpStart}{\ypStart}}
            {\pgfplotspointaxisdirectionxy{0.01}{0}}
            {\pgfplotspointaxisdirectionxy{0}{0.01}}
        };
        \draw[fill=black,draw=black] \pgfextra{
            \pgfpathellipse{\pgfplotspointaxisxy{\xpMiddle}{\ypMiddle}}
            {\pgfplotspointaxisdirectionxy{0.01}{0}}
            {\pgfplotspointaxisdirectionxy{0}{0.01}}
        };
        \draw[fill=black!20!red,draw=black!20!red] \pgfextra{
            \pgfpathellipse{\pgfplotspointaxisxy{\xpGoal}{\ypGoal}}
            {\pgfplotspointaxisdirectionxy{0.01}{0}}
            {\pgfplotspointaxisdirectionxy{0}{0.01}}
        };
        \draw[fill=blue!80!black, draw=blue!80!black] \pgfextra{
            \pgfpathellipse{\pgfplotspointaxisxy{0.3}{0.0}}
            {\pgfplotspointaxisdirectionxy{0.05}{0}}
            {\pgfplotspointaxisdirectionxy{0}{0.05}}
        };
        \draw[fill=blue!80!black, draw=blue!80!black] \pgfextra{
            \pgfpathellipse{\pgfplotspointaxisxy{0.0}{0.4}}
            {\pgfplotspointaxisdirectionxy{0.05}{0}}
            {\pgfplotspointaxisdirectionxy{0}{0.05}}
        };
        \draw[fill=blue!80!black, draw=blue!80!black] \pgfextra{
            \pgfpathellipse{\pgfplotspointaxisxy{0.2}{0.2}}
            {\pgfplotspointaxisdirectionxy{0.05}{0}}
            {\pgfplotspointaxisdirectionxy{0}{0.05}}
        };
        \addplot[
            black!20!red,
            thick,
        ] table [
            col sep=comma,
            x={x_slider},
            y={y_slider},
        ] {\table};
        \addplot[
            dashed,
            thick,
            black!20!red,
        ] table [
            col sep=comma,
            x={x_pusher},
            y={y_pusher},
        ] {\table};
    \end{axis}
    \pgfplotstableread[col sep=comma, header=true]{data/tracking_ellipse_without_disturbance.csv}{\table}
    \pgfplotstablegetrowsof{\table} 
    \pgfmathsetmacro{\Nrows}{\pgfplotsretval-1}
    \pgfplotstablegetelem{\idxStart}{x_opt}\of\table
    \pgfmathsetmacro{\xStart}{\pgfplotsretval}
    \pgfplotstablegetelem{\idxStart}{y_opt}\of\table
    \pgfmathsetmacro{\yStart}{\pgfplotsretval}
    \pgfplotstablegetelem{\idxStart}{theta_opt}\of\table
    \pgfmathsetmacro{\thetaStart}{\pgfplotsretval}
    \pgfplotstablegetelem{\idxStart}{x_pusher}\of\table
    \pgfmathsetmacro{\xpStart}{\pgfplotsretval}
    \pgfplotstablegetelem{\idxStart}{y_pusher}\of\table
    \pgfmathsetmacro{\ypStart}{\pgfplotsretval}
    \begin{axis}[
        name=trackingEllipseWithoutDist,
        clip=false,
        at={($(eight.north east)+(0.3cm,0)$)},
        anchor=north west,
        height=6.7cm,
        xmin=-0.46, xmax=0.42,
        ymin=-0.80, ymax=-0.37,
        ytick pos=right,
        ylabel={$\mathrm{y}\;[\si{\meter}]$},
        xticklabels={,,},
        axis equal image,
    ]
        \coordinate (coorStart) at (axis cs:\xStart,\yStart);
        \coordinate (coorpStart) at (axis cs:\xpStart,\ypStart);
        \pic {rectStart={coorStart,deg(\thetaStart),0.09,0.198}};
        \draw[fill=black!50!green,draw=black!50!green] \pgfextra{
            \pgfpathellipse{\pgfplotspointaxisxy{\xpStart}{\ypStart}}
            {\pgfplotspointaxisdirectionxy{0.01}{0}}
            {\pgfplotspointaxisdirectionxy{0}{0.01}}
        };
        \pgfplotsextra{
            \foreach \i in {60,120,170}{
                \pgfplotstablegetelem{\i}{x_opt}\of\table
                \pgfmathsetmacro{\xCoor}{\pgfplotsretval}
                \pgfplotstablegetelem{\i}{y_opt}\of\table
                \pgfmathsetmacro{\yCoor}{\pgfplotsretval}
                \pgfplotstablegetelem{\i}{theta_opt}\of\table
                \pgfmathsetmacro{\theta}{\pgfplotsretval};
                \pgfplotstablegetelem{\i}{x_pusher}\of\table
                \pgfmathsetmacro{\xp}{\pgfplotsretval};
                \pgfplotstablegetelem{\i}{y_pusher}\of\table
                \pgfmathsetmacro{\yp}{\pgfplotsretval};
                \coordinate (coor) at (axis cs:\xCoor,\yCoor);
                \coordinate (coorp) at (axis cs:\xp,\yp);
                \pic {rectObj={coor,deg(\theta),0.09,0.198}};
                \draw[fill=black,draw=black] \pgfextra{
                    \pgfpathellipse{\pgfplotspointaxisxy{\xp}{\yp}}
                    {\pgfplotspointaxisdirectionxy{0.008}{0}}
                    {\pgfplotspointaxisdirectionxy{0}{0.008}}
                };
            }
        }
        \addplot[
            black!20!red,
            dashdotted,
            thick,
        ] table [
            col sep=comma,
            x={x_nom},
            y={y_nom},
        ] {\table};
        \addplot[
            black,
        ] table [
            col sep=comma,
            x={x_opt},
            y={y_opt},
        ] {\table};
        \addplot[
            dashed,
            black,
        ] table [
            col sep=comma,
            x={x_pusher},
            y={y_pusher},
        ] {\table};
    \end{axis}
    \pgfplotstableread[col sep=comma, header=true]{data/tracking_ellipse_with_disturbance.csv}{\table}
    \pgfplotstablegetrowsof{\table} 
    \pgfmathsetmacro{\Nrows}{\pgfplotsretval-1}
    \pgfplotstablegetelem{\idxStart}{x_opt}\of\table
    \pgfmathsetmacro{\xStart}{\pgfplotsretval}
    \pgfplotstablegetelem{\idxStart}{y_opt}\of\table
    \pgfmathsetmacro{\yStart}{\pgfplotsretval}
    \pgfplotstablegetelem{\idxStart}{theta_opt}\of\table
    \pgfmathsetmacro{\thetaStart}{\pgfplotsretval}
    \pgfplotstablegetelem{\idxStart}{x_pusher}\of\table
    \pgfmathsetmacro{\xpStart}{\pgfplotsretval}
    \pgfplotstablegetelem{\idxStart}{y_pusher}\of\table
    \pgfmathsetmacro{\ypStart}{\pgfplotsretval}
    \begin{axis}[
        name=trackingEllipseWithDist,
        clip=false,
        at={($(trackingEllipseWithoutDist.south east)+(0,-0.27cm)$)},
        anchor=north east,
        height=6.7cm,
        xmin=-0.46, xmax=0.42,
        ymin=-0.80, ymax=-0.37,
        ytick pos=right,
        xticklabels={,,},
        ylabel={$\mathrm{y}\;[\si{\meter}]$},
        axis equal image,
    ]
        \coordinate (coorStart) at (axis cs:\xStart,\yStart);
        \coordinate (coorpStart) at (axis cs:\xpStart,\ypStart);
        \pic {rectStart={coorStart,deg(\thetaStart),0.09,0.198}};
        \draw[fill=black!50!green,draw=black!50!green] \pgfextra{
            \pgfpathellipse{\pgfplotspointaxisxy{\xpStart}{\ypStart}}
            {\pgfplotspointaxisdirectionxy{0.01}{0}}
            {\pgfplotspointaxisdirectionxy{0}{0.01}}
        };
        \pgfplotsextra{
            \foreach \i in {80,147,200}{
                \pgfmathsetmacro{\iOffset}{\i-1}
                \pgfplotstablegetelem{\i}{x_opt}\of\table
                \pgfmathsetmacro{\xCoor}{\pgfplotsretval}
                \pgfplotstablegetelem{\i}{y_opt}\of\table
                \pgfmathsetmacro{\yCoor}{\pgfplotsretval}
                \pgfplotstablegetelem{\i}{theta_opt}\of\table
                \pgfmathsetmacro{\theta}{\pgfplotsretval};
                \pgfplotstablegetelem{\iOffset}{x_pusher}\of\table
                \pgfmathsetmacro{\xp}{\pgfplotsretval};
                \pgfplotstablegetelem{\iOffset}{y_pusher}\of\table
                \pgfmathsetmacro{\yp}{\pgfplotsretval};
                \coordinate (coor) at (axis cs:\xCoor,\yCoor);
                \coordinate (coorp) at (axis cs:\xp,\yp);
                \pic {rectObj={coor,deg(\theta),0.09,0.198}};
                \draw[fill=black,draw=black] \pgfextra{
                    \pgfpathellipse{\pgfplotspointaxisxy{\xp}{\yp}}
                    {\pgfplotspointaxisdirectionxy{0.008}{0}}
                    {\pgfplotspointaxisdirectionxy{0}{0.008}}
                };
            }
        }
        \pgfplotsextra{
            \foreach \i in {46,61,76,160,174,196}{
                \pgfplotstablegetelem{\i}{x_opt}\of\table
                \pgfmathsetmacro{\xCoor}{\pgfplotsretval}
                \pgfplotstablegetelem{\i}{y_opt}\of\table
                \pgfmathsetmacro{\yCoor}{\pgfplotsretval}
                \coordinate (coor) at (axis cs:\xCoor,\yCoor);
                \draw[color=black!20!red, mark size=4.0pt, thick] plot[mark=x] coordinates{(coor)} node {};
            }
        }
        \addplot[
            thick,
            dashdotted,
            black!20!red,
        ] table [
            col sep=comma,
            x={x_nom},
            y={y_nom},
        ] {\table};
        \addplot[
            black,
        ] table [
            col sep=comma,
            x={x_opt},
            y={y_opt},
        ] {\table};
        \addplot[
            dashed,
            black,
        ] table [
            col sep=comma,
            x={x_pusher},
            y={y_pusher},
        ] {\table};
    \end{axis}
    \def\idxMiddle{60}
    \def\xl{0.09}
    \def\yl{0.198}
    \def\rObs{0.065}
    \def\rp{0.008}
    \pgfplotstableread[col sep=comma, header=true]{data/planning_with_obstacle.csv}{\table}
    \pgfplotstablegetrowsof{\table} 
    \pgfmathsetmacro{\Nrows}{\pgfplotsretval-1}
    \pgfmathsetmacro{\NrowsOffset}{\Nrows-2}
    \pgfplotstablegetelem{\idxStart}{x_opt}\of\table
    \pgfmathsetmacro{\xStart}{\pgfplotsretval}
    \pgfplotstablegetelem{\idxStart}{y_opt}\of\table
    \pgfmathsetmacro{\yStart}{\pgfplotsretval}
    \pgfplotstablegetelem{\idxStart}{theta_opt}\of\table
    \pgfmathsetmacro{\thetaStart}{\pgfplotsretval}
    \pgfplotstablegetelem{\idxStart}{x_pusher}\of\table
    \pgfmathsetmacro{\xpStart}{\pgfplotsretval}
    \pgfplotstablegetelem{\idxStart}{y_pusher}\of\table
    \pgfmathsetmacro{\ypStart}{\pgfplotsretval}
    \pgfplotstablegetelem{\idxStart}{x_obs}\of\table
    \pgfmathsetmacro{\xObs}{\pgfplotsretval}
    \pgfplotstablegetelem{\idxStart}{y_obs}\of\table
    \pgfmathsetmacro{\yObs}{\pgfplotsretval}
    \pgfplotstablegetelem{\NrowsOffset}{x_nom}\of\table
    \pgfmathsetmacro{\xGoal}{\pgfplotsretval}
    \pgfplotstablegetelem{\NrowsOffset}{y_nom}\of\table
    \pgfmathsetmacro{\yGoal}{\pgfplotsretval}
    \begin{axis}[
        name=planningExp,
        clip=false,
        at={($(planningWithObs.south east)+(0.3cm,0)$)},
        anchor=south west,
        height=6.7cm,
        xmin=-0.46, xmax=0.42,
        ymin=-0.75, ymax=-0.30,
        ytick pos=right,
        xlabel={$\mathrm{x}\;[\si{\meter}]$},
        ylabel={$\mathrm{y}\;[\si{\meter}]$},
        axis equal image,
    ]
        \coordinate (coorStart) at (axis cs:\xStart,\yStart);
        \coordinate (coorpStart) at (axis cs:\xpStart,\ypStart);
        \coordinate (coorGoal) at (axis cs:\xGoal,\yGoal);
        \pic {rectStart={coorStart,deg(\thetaStart),\xl,\yl}};
        \pic {rectGoal={coorGoal,180,\xl,\yl}};
        \draw[fill=black!50!green,draw=black!50!green] \pgfextra{
            \pgfpathellipse{\pgfplotspointaxisxy{\xpStart}{\ypStart}}
            {\pgfplotspointaxisdirectionxy{0.01}{0}}
            {\pgfplotspointaxisdirectionxy{0}{0.01}}
        };
        \draw[fill=blue!80!black,draw=blue!80!black] \pgfextra{
            \pgfpathellipse{\pgfplotspointaxisxy{\xObs}{\yObs}}
            {\pgfplotspointaxisdirectionxy{\rObs}{0}}
            {\pgfplotspointaxisdirectionxy{0}{\rObs}}
        };
        \pgfplotsextra{
            \foreach \i in {72, 90}{
                \pgfmathsetmacro{\iOffset}{\i-2}
                \pgfplotstablegetelem{\i}{x_opt}\of\table
                \pgfmathsetmacro{\xCoor}{\pgfplotsretval}
                \pgfplotstablegetelem{\i}{y_opt}\of\table
                \pgfmathsetmacro{\yCoor}{\pgfplotsretval}
                \pgfplotstablegetelem{\i}{theta_opt}\of\table
                \pgfmathsetmacro{\theta}{\pgfplotsretval};
                \pgfplotstablegetelem{\iOffset}{x_pusher}\of\table
                \pgfmathsetmacro{\xp}{\pgfplotsretval};
                \pgfplotstablegetelem{\iOffset}{y_pusher}\of\table
                \pgfmathsetmacro{\yp}{\pgfplotsretval};
                \coordinate (coor) at (axis cs:\xCoor,\yCoor);
                \coordinate (coorp) at (axis cs:\xp,\yp);
                \pic {rectObj={coor,deg(\theta),\xl,\yl}};
                \draw[fill=black,draw=black] \pgfextra{
                    \pgfpathellipse{\pgfplotspointaxisxy{\xp}{\yp}}
                    {\pgfplotspointaxisdirectionxy{\rp}{0}}
                    {\pgfplotspointaxisdirectionxy{0}{\rp}}
                };
            }
        }
        \addplot[
            thick,
            black!20!red,
        ] table [
            col sep=comma,
            x={x_nom},
            y={y_nom},
            restrict expr to domain={\thisrow{idx}}{0:\NrowsOffset},
        ] {\table};
        \addplot[
            thick,
            black!20!red,
            dashed,
        ] table [
            col sep=comma,
            x expr={\thisrow{x_nom}-((\xl/2.)+\rp)*cos(deg(\thisrow{theta_nom}))+(\xl/2.)*tan(deg(\thisrow{psi_nom}))*sin(deg(\thisrow{theta_nom}))},
            y expr={\thisrow{y_nom}-((\xl/2.)+\rp)*sin(deg(\thisrow{theta_nom}))-(\xl/2.)*tan(deg(\thisrow{psi_nom}))*cos(deg(\thisrow{theta_nom}))},
            restrict expr to domain={\thisrow{idx}}{0:\NrowsOffset},
        ] {\table};
        \addplot[
            black,
        ] table [
            col sep=comma,
            x={x_opt},
            y={y_opt},
        ] {\table};
        \addplot[
            dashed,
            black,
        ] table [
            col sep=comma,
            x={x_pusher},
            y={y_pusher},
        ] {\table};
    \end{axis}
    \node[above=0.4cm, anchor=south] (legend) at ($(circle.north west)!0.5!(trackingEllipseWithoutDist.north east)$) {\pgfplotslegendfromname{commonLegendTraj}};
    \node (labela) at ($(circle.north west)$) [left=0.2cm, text width=2em, anchor=south west, align=center] {\subcaption{}\label{fig:tracking_circle_traj}};
    \node (labelb) at ($(eight.north west)$) [left=0.2cm, text width=2em, anchor=south west, align=center] {\subcaption{}\label{fig:tracking_eight_traj}};
    \node (labelc) at ($(planningWithoutObs.north west)$) [left=0.2cm, text width=2em, anchor=south west, align=center] {\subcaption{}\label{fig:plannning_without_obstacles}};
    \node (labeld) at ($(planningWithObs.north west)$) [left=0.2cm, text width=2em, anchor=south west, align=center] {\subcaption{}\label{fig:planning_with_obstacles}};
    \node (labele) at ($(trackingEllipseWithoutDist.north west)$) [left=0.2cm, text width=2em, anchor=south west, align=center] {\subcaption{}\label{fig:tracking_ellipse_without_disturbances}};
    \node (labelf) at ($(trackingEllipseWithDist.north west)$) [left=0.2cm, text width=2em, anchor=south west, align=center] {\subcaption{}\label{fig:tracking_ellipse_with_disturbances}};
    \node (labelg) at ($(planningExp.north west)$) [left=0.2cm, text width=2em, anchor=south west, align=center] {\subcaption{}\label{fig:planning_with_obstacle_experiment}};
    \draw[very thick] ($(eight.north east)+(1.5mm,4mm)$) -- ($(planningWithObs.south east)+(1.5mm,-8mm)$);
    \node (arrowRight) at ($(eight.north east)+(1.5mm,4mm)$) [below right, anchor=north west] {$\longrightarrow$};
    \node (arrowLeft) at ($(eight.north east)+(1.5mm,4mm)$) [below left, anchor=north east] {$\longleftarrow$};
    \node (text1) at ($(circle.north west)!0.5!(eight.north east)$) [above] {\textbf{numerical experiment results}};
    \node (text2) at ($(trackingEllipseWithoutDist.north west)!0.5!(trackingEllipseWithoutDist.north east)$) [above] {\textbf{robot experiment results}};
\end{tikzpicture}

%% file: Tikz/fig_results_plots.tex
\begin{tikzpicture}
    \pgfplotstableread[col sep=comma, header=true]{data/tracking_circle_cc_action.csv}{\tableCCMPC}
    \pgfplotstableread[col sep=comma, header=true]{data/tracking_circle_mi_action.csv}{\tableMIMPC}
    \pgfplotstableread[col sep=comma, header=true]{data/planning_times_cc_mean.csv}{\tableCCPlanning}
    \pgfplotstableread[col sep=comma, header=true]{data/planning_times_mi_mean.csv}{\tableMIPlanning}
    \pgfplotstableread[col sep=comma]{data/error_with_dist_mi.csv}{\tableMIDist}%
    \pgfplotstableread[col sep=comma]{data/error_with_dist_cc.csv}{\tableCCDist}%
    \pgfplotstableread[col sep=comma]{data/time_horizon_comparison_mi.csv}{\tableMITH}%
    \pgfplotstableread[col sep=comma]{data/time_horizon_comparison_cc.csv}{\tableCCTH}%
    \begin{axis}[
        name=solvingTimeMPC,
        ymin = 0.0, ymax=100,
        xlabel={Time$\;[\si{\second}]$},
        ylabel={Duration$\;[\si{\milli\second}]$},
        grid=both,
        enlargelimits=false,
        width=0.5\textwidth,
        height=3.0cm,
        legend style={column sep=2mm},
        legend to name={commonLegendRes},
        legend columns = 5,
    ]
        \addlegendimage{color=blue,mark=x}
        \addlegendentry{MIQP}
        \addlegendimage{color=blue,mark=o}
        \addlegendentry{MINLP}
        \addlegendimage{color=black,mark=+}
        \addlegendentry{MPCC}
        \addlegendimage{dashed,very thick,color=red}
        \addlegendentry{$50\;\si{\hertz}$}
        \addlegendimage{mark=x,only marks,black!20!red,mark size=3.0pt,thick}
        \addlegendentry{Disturbance}
        \addplot[
            mark=x,
            only marks,
            blue,
        ] table [
            col sep=comma,
            x={time},
            y expr={\thisrow{comp_time}*1000},
        ] {\tableMIMPC};
        \addplot[
            mark=+,
            only marks,
            black,
        ] table [
            col sep=comma,
            x={time},
            y expr={\thisrow{comp_time}*1000},
        ] {\tableCCMPC};
        \draw[red, dashed, very thick] (\pgfkeysvalueof{/pgfplots/xmin}, 20) -- (\pgfkeysvalueof{/pgfplots/xmax}, 20);
        \coordinate (coorb) at (axis cs:5,20);
        \draw[color=black!20!red, mark size=6.0pt, ultra thick] plot[mark=x] coordinates{(coorb)} node {};
    \end{axis}
    \begin{axis}[
      name=errorMPC,
      at={($(solvingTimeMPC.north east)+(1.5cm,0.)$)},
      anchor=north west,
      ymode=log,
      log basis y={10},
      xlabel={Angular disturbance$\;[\si{\radian\per\second}]$},
      ylabel={Error$\;[\si{\milli\meter}]$},
      grid=major,
      enlargelimits=false,
      width=0.5\textwidth,
      height=3.0cm,
      ]
      \addplot[mark=x, thick, blue] table [x={ang_dist}, y={err_median}] {\tableMIDist};
      \addplot[mark=+, thick, black] table [x={ang_dist}, y={err_median}] {\tableCCDist};
      \addplot[name path=mi_lower,draw=none] table [x={ang_dist},y={err_lower}] {\tableMIDist};
      \addplot[name path=mi_upper,draw=none] table [x={ang_dist},y={err_upper}] {\tableMIDist};
      \addplot[fill=blue, opacity=0.2] fill between [of=mi_lower and mi_upper];
      \addplot[name path=cc_lower,draw=none] table [x={ang_dist},y={err_lower}] {\tableCCDist};
      \addplot[name path=cc_upper,draw=none] table [x={ang_dist},y={err_upper}] {\tableCCDist};
      \addplot[fill=black, opacity=0.2] fill between [of=cc_lower and cc_upper];
    \end{axis}
    \begin{axis}[
      name=timeHorizonMPC,
      at={($(solvingTimeMPC.south east)+(0,-0.9cm)$)},
      anchor=north east,
      xlabel={Horizon$\;[N]$},
      ylabel={Duration$\;[\si{\milli\second}]$},
      grid=both,
      enlargelimits=false,
      width=0.5\textwidth,
      height=3.0cm,
      ]
      \addplot[mark=x, thick, blue] table [x={th}, y={comp_time_median}] {\tableMITH};
      \addplot[mark=+, thick, black] table [x={th}, y={comp_time_median}] {\tableCCTH};
      \addplot[name path=mi_lower,draw=none] table [x={th},y={comp_time_lower}] {\tableMITH};
      \addplot[name path=mi_upper,draw=none] table [x={th},y={comp_time_upper}] {\tableMITH};
      \addplot[fill=blue, opacity=0.2] fill between [of=mi_lower and mi_upper];
      \addplot[name path=cc_lower,draw=none] table [x={th},y={comp_time_lower}] {\tableCCTH};
      \addplot[name path=cc_upper,draw=none] table [x={th},y={comp_time_upper}] {\tableCCTH};
      \addplot[fill=black, opacity=0.2] fill between [of=cc_lower and cc_upper];
      \draw[red, dashed, very thick] (\pgfkeysvalueof{/pgfplots/xmin}, 20) -- (\pgfkeysvalueof{/pgfplots/xmax}, 20);
    \end{axis}
    \begin{axis}[
        name=solvingTimePlanning,
        at={($(errorMPC.south east)+(0,-0.9cm)$)},
        anchor=north east,
        xlabel={Time horizon$\;[\si{\second}]$},
        ylabel={Duration$\;[\si{\second}]$},
        grid=both,
        enlargelimits=false,
        width=0.5\textwidth,
        height=3.0cm,
    ]
        \addplot[
            mark=o,
            blue,
        ] table [
            col sep=comma,
            x={horizon},
            y={mean_time},
        ] {\tableMIPlanning};
        \addplot[
            mark=+,
            black,
        ] table [
            col sep=comma,
            x={horizon},
            y={mean_time},
        ] {\tableCCPlanning};
    \end{axis}
    \node[above=0.0cm, anchor=south] (legend) at ($(solvingTimeMPC.north west)!0.5!(errorMPC.north east)$) {\pgfplotslegendfromname{commonLegendRes}};
    \node[rectangle,fill=white,text width=0.5em, below=0.1cm, anchor=north] (labelaREC) at ($(solvingTimeMPC.north)$) {};
    \node (labela) at ($(solvingTimeMPC.north)$) [text width=1em,anchor=south,align=center] {\subcaption{}\label{fig:tracking_circle_comp_time}};
    \node[rectangle,fill=white,text width=0.5em, below=0.1cm, anchor=north] (labelcREC) at ($(errorMPC.north)$) {};
    \node (labelb) at ($(errorMPC.north)$) [text width=1em,anchor=south,align=center] {\subcaption{}\label{fig:tracking_circle_disturbance_error}};
    \node[rectangle,fill=white,text width=0.5em, below=0.1cm, anchor=north] (labeldREC) at ($(timeHorizonMPC.north)$) {};
    \node (labelc) at ($(timeHorizonMPC.north)$) [text width=1em,anchor=south,align=center] {\subcaption{}\label{fig:tracking_circle_time_horizon_comp_time}};
    \node[rectangle,fill=white,text width=0.5em, below=0.1cm, anchor=north] (labelbREC) at ($(solvingTimePlanning.north)$) {};
    \node (labeld) at ($(solvingTimePlanning.north)$) [text width=1em,anchor=south,align=center] {\subcaption{}\label{fig:planning_comp_time}};
\end{tikzpicture}

%% file: sections/conclusion.tex
\paraDraft{Summary}
This paper presents a~\acrfull*{mpcc} formulation for generating hybrid trajectories for planar pushing manipulation tasks.
We show that this formulation is able to:
\begin{enumerate*}[label=(\roman*)]
\item plan trajectories that exploit both sticking and sliding contact modes in scenarios with obstacles, and 
\item track nominal trajectories, using an~\acrshort*{mpc} loop, under external disturbances and model uncertainties.
\end{enumerate*}
We compare the proposed~\acrshort*{mpcc} formulation with the mixed integer alternatives,~\textit{i.e.} with~\acrshort*{minlp} for planning trajectories and~\acrshort*{miqp} for control, showing that our formulation is 
\begin{enumerate*}[label=\roman*)]
  \item computationally faster for planning problems,
  \item computationally more reliable across different scenarios,
  \item and better able to track challenging nominal trajectories and recover from disturbances.
\end{enumerate*}
Furthermore, we tested our planner and~\acrshort*{mpc} implementation on a KUKA LWR robot setup without any model identification, which demonstrates the reliability of the proposed controller.

\paraDraft{Discussion and future work}
This work demonstrates the potential of using complementarity constraints within an~\acrshort*{mpc} for planar manipulation tasks involving contacts with friction.
It remains to study the scalability of such approach with respect to number of contacts and different tasks, such as pivoting or non quasi-static scenarios.
For future work, we will focus on improving the tracking performance via model selection and identification.
Additionally, we aim to improve the~\acrshort*{mpc} computation times towards increasing the time horizon, making it less short-sighted, and enabling its deployment in environments with highly dynamic obstacles.